\documentclass[lettersize,journal]{IEEEtran}
\usepackage{amsmath,amsfonts}
\usepackage{algorithmic}
\usepackage{algorithm}
\usepackage{array}
\usepackage{diagbox}
\usepackage[caption=false,font=normalsize,labelfont=sf,textfont=sf]{subfig}
\usepackage{amsthm,cite,url,graphicx,booktabs,lipsum,color,bm,caption,soul}

\usepackage{textcomp}
\usepackage{marvosym}
\usepackage{tikz}
\usepackage{stfloats}
\usepackage[utf8]{inputenc}
\usepackage{textgreek}
\usepackage{makecell}
\usepackage{url}
\usepackage{hyperref}
\usepackage{colortbl}
\usepackage{xcolor}
\usepackage{verbatim}
\usepackage{graphicx}
\usepackage{subfig}
\usepackage{graphicx}
\usepackage{subfig}
\usepackage{cite}
\usepackage{threeparttable}
\usepackage{booktabs}
\usepackage{multirow}
\usepackage{adjustbox}
\usepackage{graphicx}
\usepackage{subcaption}
\usepackage{float}
\usepackage[utf8]{inputenc}
\usepackage{amsmath, amssymb, bm}
\usepackage[T1]{fontenc}
\usepackage{bm}
\usepackage{makecell} 
\begin{document}

\title{Variational Mixture of Graph Neural Experts for Alzheimer's Disease Recognition across Frequency Bands in EEG Brain Networks}

\author{Jun-En Ding, Anna Zilverstand, Shihao Yang, Albert Chih-Chieh Yang, and Feng Liu\textsuperscript{\Letter}
\thanks{J. Ding is with the Department of Systems Engineering, Stevens Institute of Technology, Hoboken, NJ 07030, USA.}
\thanks{S. Yang and F. Liu are with the Department of Industrial and Systems Engineering, Rutgers University, Piscataway, NJ, USA.}
\thanks{A. Zilverstand is with the Department of Psychiatry and Behavioral Sciences, University of Minnesota, Minneapolis, MN 55414, USA.}
\thanks{A. C.-C. Yang is with the Institute of Brain Science, College of Medicine, National Yang-Ming Chiao Tung University, Taipei City, Taiwan.}
\thanks{\textsuperscript{\Letter}Corresponding author: F. Liu, Department of Industrial \& Systems Engineering, Rutgers University, Piscataway, NJ, USA (Email: feng.liu@rutgers.edu).} 
}
\maketitle

\begin{abstract}
    Dementia disorders such as Alzheimer's disease (AD) and frontotemporal dementia (FTD) exhibit overlapping electrophysiological signatures in electroencephalography (EEG) that challenge accurate diagnosis. Existing EEG-based methods are limited by full-band frequency analysis, which hinders precise differentiation of dementia subtypes and severity stages. To address this limitation, we propose a Variational Mixture of Graph Neural Experts (VMoGE) framework that integrates multi-band EEG analysis with variational graph neural networks and a mixture-of-experts architecture. Each expert specializes in a specific EEG frequency band and models brain connectivity using a Gaussian Markov Random Field prior, while a variational gating mechanism adaptively integrates expert outputs. This design enables the model to learn frequency-specific brain network representations while modeling latent uncertainty through variational inference. Experimental results on two EEG dementia datasets show that VMoGE achieves strong performance, with an area under the curve (AUC) of 0.89 for healthy controls (HC) vs. AD classification in the main comparison and competitive results across dementia subtyping and Clinical Dementia Rating (CDR) staging tasks. Clinically, VMoGE offers three key translational values: the expert gating weights correlate with Mini-Mental State Examination (MMSE) scores and CDR severity, slow-wave $\delta / \theta$-band contributions are associated with AD-related EEG slowing and disease progression, and spatially localized activation maps reveal posterior $\theta$/$\alpha$-band alterations and region-specific $\beta$-band changes, providing neurophysiologically interpretable patterns aligned with known AD neuropathology.

\end{abstract}

\newenvironment{IEEEnote}
{\par\noindent\hspace*{1em}\textbf{\textit{Note to Practitioners}}---\ignorespaces}
{\par}

\begin{IEEEnote}
\textbf{The clinical diagnosis of dementia often relies on physicians' experience-based interpretation of EEG waveforms and clinical indicators. Many existing studies lack a comprehensive evaluation that simultaneously considers model performance on both EEG biomarkers and clinical indicators for Alzheimer's disease assessment. This study proposes the variational mixture of graph neural experts (VMoGE), a mixture of graph experts integrating multi-band feature analysis with variational inference. VMoGE automatically learns brain network activity patterns across different frequency bands and visualizes disease-associated brain regions and spectral features. Additionally, VMoGE provides automated dementia biomarker identification, with each expert module corresponding to specific frequency bands. Weight variations correlate with clinical scales (e.g., MMSE), age, and disease progression, thereby enhancing the efficiency and objectivity of detecting early-stage AD biomarker changes and identifying high-risk cases.}
\end{IEEEnote}

\section{Introduction}

Alzheimer's disease (AD) is the most common cause of dementia worldwide. In the United States, an estimated 6.9 million individuals aged 65 years and older were living with Alzheimer's dementia in 2024~\cite{alzheimers2025facts}.
Frontotemporal dementia (FTD) is a major cause of young-onset dementia, typically presenting between 35 and 75 years of age. It is characterized by progressive degeneration of the frontal and temporal lobes, leading to early changes in behavior, personality, and language. Common manifestations include disinhibition, emotional blunting, stereotyped or repetitive behaviors, and altered eating habits~\cite{weder2007frontotemporal,bathgate2001behaviour}.

Traditional diagnostic workflows rely on neuropsychological assessments, such as the Mini-Mental State Examination (MMSE) and Montreal Cognitive Assessment (MoCA); structural neuroimaging, including magnetic resonance imaging (MRI) and computed tomography (CT); and invasive biomarker analysis, such as cerebrospinal fluid   A$\beta$42/tau ratios and amyloid positron emission tomography (PET)~\cite{hansson2019advantages}.
Despite their diagnostic value, these methods face limitations that restrict widespread adoption. Neuropsychological tests such as the MMSE suffer from ceiling effects in early-stage dementia and are vulnerable to practice effects~\cite{gluhm2013cognitive,o2016screening,d2025mmse}. Invasive procedures such as lumbar puncture for CSF collection reduce patient compliance and preclude routine longitudinal monitoring, particularly in elderly populations at elevated procedural risk~\cite{baldaranov2023safety,vandevrede2025clinical}. Neuroimaging biomarkers demand costly equipment and specialized expertise, rendering them largely inaccessible in resource-constrained settings~\cite{vrahatis2023revolutionizing,mcmahon2003cost,mattsson2024plasma}.  The Electroencephalography (EEG) is a cost-effective neuroimaging modality, which effectively reflecting cortical activities through brain signals \cite{He2016ImagingMRI}.  Several studies have demonstrated that AD is associated with disrupted
brain connectivity, reduced neural flexibility, and abnormal EEG state
transitions, reflecting impaired large-scale network coordination and functional
degeneration~\cite{razavi2025brain, sandonis2025disrupted}.

Recent research findings indicate that analyzing time-frequency differences across bands EEG can potentially identify AD related EEG biomarkers. For example, both AD and FTD typically exhibit spectral slowing, characterized by increased delta/theta power and reduced $\alpha$/$\beta$ power, which correlates with cognitive decline~\cite{lin2021differences,jiao2023neural}. Early-onset AD often demonstrates widespread delta enhancement, whereas FTD shows more region-specific abnormalities and distinct microstate alterations~\cite{lin2021differences}. Power ratios such as theta/alpha and theta/beta, along with gamma band changes, have been proposed as sensitive markers for prodromal and clinical AD~\cite{chetty2024eeg}.

The success of deep learning has greatly enhanced feature extraction from multi-band EEG signals, such as convolutional neural networks (CNNs), long short-term memory networks (LSTMs), and hybrid architectures, which can automatically learn spatiotemporal features across frequency bands~\cite{khan2025explainable,rezaee2025diagnose}. Specifically, Transformer architectures have demonstrated significant potential for EEG-based AD classification~\cite{cansiz2025olfactory,latif2025deep,cseker2025investigating}. The field has progressed to convolutional-Transformer hybrid architectures such as CEEDNet and DICE-Net~\cite{miltiadous2023dice} to large-scale foundation models with cross-dataset pretraining, with models like LEAD achieving classification accuracies exceeding 93\% and establishing Transformers as a pivotal direction in EEG-based dementia diagnosis~\cite{wang2025lead}.


Building on these advancements, and unlike traditional EEG feature extraction approaches, more recent research has explored graph neural networks (GNNs), which have emerged as a powerful architecture for modeling EEG brain connectivity and capturing complex spatial, spectral, and temporal dependencies. Furthermore, multi-path, multi-frequency, and attention-based GNN architectures demonstrate superior performance in distinguishing dementia-related disorders, offering promising tools for accurate and interpretable diagnosis~\cite{zhang2025dual,xu2025multi,yang2026large,yang2024rejuvenating}.

Research on GNNs integrated with variational inference has demonstrated strong potential in handling graph data with inherent uncertainty and noise. Extending to spatiotemporal settings, a new framework by Liang \textit{et al.} leverages diffusion processes and variational inference to uncover dynamic causal relationships with enhanced accuracy and interpretability~\cite{liang2024dynamic}. Although these approaches often consider standard Gaussian priors as prior distributions, variational approximations remain insufficient to fully capture complex distributions, particularly brain signal networks with graph structures.

Recently,  Mixture of Experts (MoE) has gained significant attention due to its key advantage of dynamic task allocation, which leverages expert specialization to significantly reduce computational costs while maintaining high performance, especially in Large Language Models (LLMs) or dynamics of multimodal fusion~\cite{wang2025dynamical}.  Furthermore, GNN integrated with MoE has significantly improved graph and node classification accuracy~\cite{shi2025mixture}. For example, GraphDIVE employs mixture-of-experts with semantic partitioning to address imbalanced graph classification by optimizing the evidence lower bound, effectively reducing bias toward majority classes~\cite{hu2021graph}. While GNNs provide powerful structural representations, MoE introduces specialized expert modules dynamically routed by gating mechanisms~\cite{shirasuna2024multi,huang2025graph}.


Nevertheless, despite substantial efforts to identify EEG-based biomarkers for AD, their clinical utility remains limited by heterogeneity in signal preprocessing procedures~\cite{del2025more}, variations in experimental protocols and data acquisition across subjects~\cite{wang2025lead,rezaei2025future}, and the increasing reliance on deep learning models for automated EEG feature extraction, which may reduce the interpretability of the resulting biomarkers for dementia diagnosis~\cite{lopes2023using,sreedhar2025fuzzy}. In addition, the clinical characterization of AD extends beyond its differentiation from other dementia subtypes, such as FTD. Disease severity and progression are commonly evaluated through cognitive performance, functional abilities, and neuropsychiatric symptoms assessed using behavioral observations and standardized clinical scales. Consequently, these methodological and clinical challenges have hindered the development of EEG biomarkers that are simultaneously effective, interpretable, objective, and straightforward to implement in clinical diagnostic applications~\cite{yang2013cognitive}.

To bridge this gap, we first develop a multi-granularity EEG feature extractor designed to learn node-level representations that capture both different-scale temporal patterns. Subsequently,  we introduce a variational mixture of graph neural experts (VMoGE) where each expert is modeled under a Gaussian Markov Random Field (GMRF) prior corresponding to a distinct EEG frequency band. By embedding structured variational inference into the mixture-of-experts paradigm, our VMoGE adaptively assigns responsibilities among experts while maintaining the intrinsic graph topology and effectively representing epistemic uncertainty, which contributes to the classification and staging of dementia. The main contributions of this paper are as follows:

\begin{itemize}
    \item  This study presents a Multi-Granularity Transformer for Node Feature Extraction (MGT-NFE) that captures multi-scale temporal patterns across four EEG frequency bands by integrating 1-D CNN modules at multiple temporal granularities with Fast Fourier Transform (FFT)-based spectral features, yielding rich frequency-specific node-level representations for graph-based dementia classification.

    \item We propose a variational mixture of graph neural experts where each expert is governed by a band-specific GMRF prior. Through a closed-form Kullback–Leibler (KL) divergence formulation, VMoGE enforces neurophysiologically meaningful spatial smoothness constraints, improving generalization under limited-sample clinical conditions.
    
    \item Through adaptive gating and evidence lower bound optimization, VMoGE dynamically assigns weights to frequency-specific experts, providing interpretable insights into EEG biomarkers, including expert weights that correlate with clinical indicators (e.g., MMSE scores, age, and disease progression) and spatial patterns aligned with neuropathological signatures of AD and FTD.

\end{itemize}

The structure of this study is as follows. Section~\ref{sec:related_work} reviews related work. Section~\ref{sec:methods} presents the problem formulation, the MGT-NFE module, GMRF prior modeling, and the variational graph convolutional encoder. Section~\ref{sec:VMoGE} details the VMoGE framework, including the gating mechanism and evidence lower bound (ELBO) optimization. Section~\ref{sec:experiments} describes the experimental setup, dataset descriptions, baseline comparisons, and ablation studies. Section~\ref{sec:noise_analysis} evaluates model robustness under varying noise conditions. Section~\ref{sec:explainable_diagnosis} provides explainable diagnosis analysis via gating weights, clinical indicators, and spatial brain activation patterns. Section~\ref{sec:discussion} discusses the frequency-specific biomarkers, spatial patterns, and neurophysiological mechanisms revealed by VMoGE, as well as its potential for multimodal extension.
	
\section{Related Work}\label{sec:related_work}


\subsection{Variational Inference Framework in EEG}

Existing variational inference methods have established a robust framework for EEG analysis, providing principled solutions for representation learning, uncertainty quantification, and modeling individual differences in brain activity patterns. For example, EEG2Vec uses a conditional variational autoencoder that learns both generative and discriminative representations from affective EEG data, achieving 68.49\% accuracy in emotion classification while generating synthetic EEG sequences that preserve low-frequency spectral characteristics~\cite{bethge2022eeg2vec}.  Building on graph-based approaches, Zhang \textit{et al.} ~\cite{zhang2020variational} proposed variational pathway reasoning, which uses random walks within brain regions to model emotion-related pathways and Bayesian processes to learn pathway importance, achieving 94.3\% accuracy on the SEED dataset with interpretable visualizations. To address individual variability, the variational instance-Adaptive Graph  combines deterministic instance-adaptive graphs with probabilistic variational graphs to capture both subject-specific dependencies and underlying uncertainties~\cite{song2021variational}. Complementing these approaches,~\cite{ofner2020balancing} explored variational predictive coding through hierarchical recurrent state-space models that balance active learning and active inference, enabling multi-level EEG prediction while maintaining biological plausibility. Moreover, the EEG graph combines a variational spatial encoder with a Gaussian temporal encoder to capture both structural brain priors and delayed cross-regional temporal dependencies~\cite{liu2024vsgt}.

\subsection{Mixture-of-Experts for Brain Network Analysis }
Recent studies have demonstrated the potential of mixture-of-experts (MoE) architectures for modeling the heterogeneity of neurophysiological and neuroimaging data, including EEG~\cite{gui2024eegmamba} and functional magnetic resonance imaging (fMRI)~\cite{zhu2025decoding}. By combining specialized expert modules with adaptive gating mechanisms, MoE frameworks can accommodate inter-subject variability and heterogeneous neural representations~\cite{yang2022eeg}. In EEG applications, expert specialization has been explored for seizure classification and cross-subject neural decoding, where different experts capture complementary signal characteristics or brain-region-specific representations~\cite{wu2025neuro,du2023mixture}.

Despite the promising ability of MoE models to capture features across distinct dynamic regions of EEG brain networks, current approaches seldom integrate more fine-grained frequency-based graph structures as prior distributions. This limitation is particularly relevant in FTD research, where electrophysiological signatures differ from those observed in AD. For instance, AD is often characterized by rostral dominance in fractal complexity, whereas FTD exhibits caudal dominance, alongside distinct alterations in slow-frequency activity~\cite{ghassemkhani2025evaluating}. In this work, we demonstrate that incorporating structure-specific graph priors that reflect such frequency-band asymmetries can substantially improve the interpretability of EEG biomarkers, enabling MoE frameworks to better distinguish between FTD and AD cohorts.

\begin{figure}
    \centering
    \includegraphics[width=0.5\textwidth]{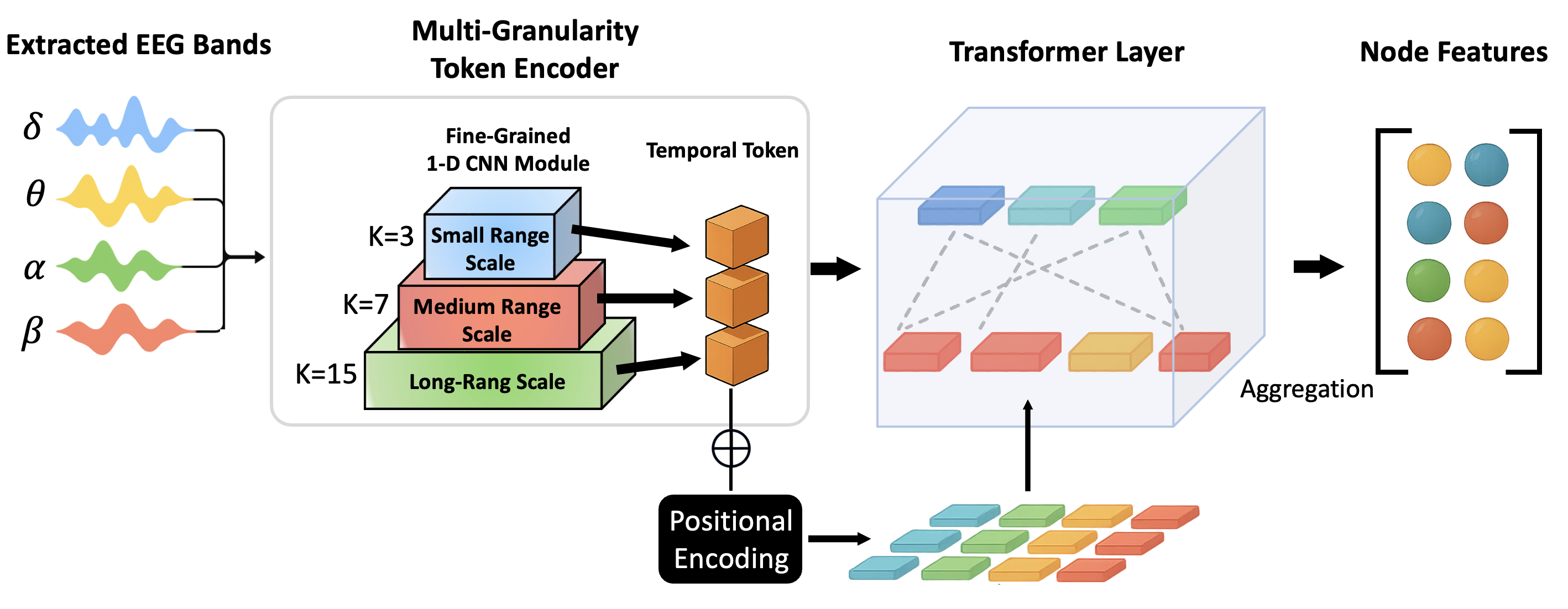}
    \caption{Diagram of MGT-NFE for node feature extraction, incorporating multi-granularity hierarchical feature extraction and spatial positional encoding at different granularities.}
    \label{fig:Transformer}
\end{figure}

\section{Methods}\label{sec:methods}


 \begin{figure*}
    \centering
    \includegraphics[width=0.95\textwidth]{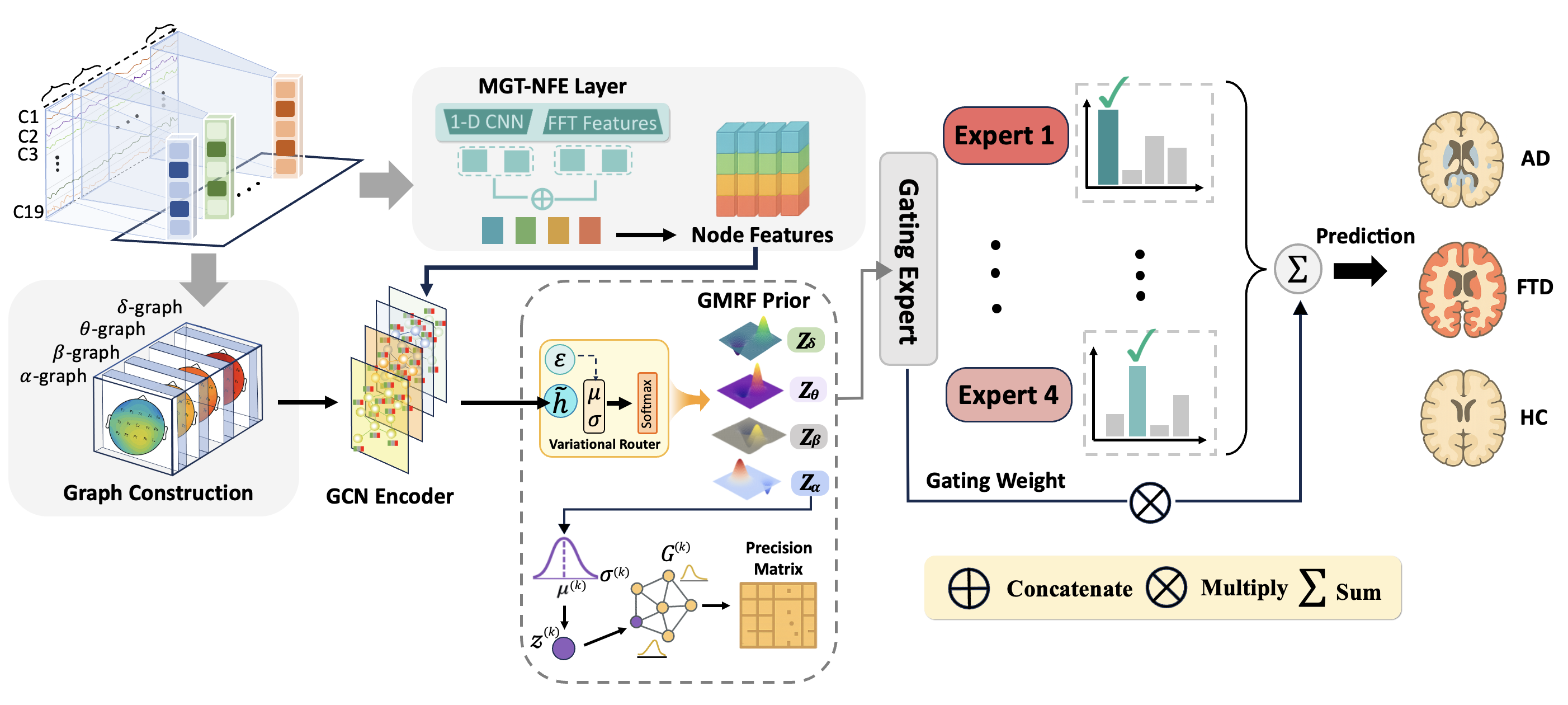}
    \caption{\textbf{Overview of the VMoGE framework for AD biomarker identification and prediction.} VMoGE framework first extracts node features from multi-channel EEG signals (channels C1–C19) by integrating spatial and frequency band features through a 1D-CNN and FFT-based MGT-NFE module. The prior graph structure is constructed using a GMRF, and a variational router models the latent distribution to capture structural correlations across multiple frequency bands. Finally, the framework performs a weighted sum of four expert networks using k-th gating probabilities to obtain the final classification result.}
    \label{fig:Fig1}
    \vspace{-2mm}
\end{figure*}

\subsection{Problem Formulation}

In this work, we implement the common feature-extraction pipeline in two datasets. The feature extraction process was implemented through a systematic epoching and spectral analysis pipeline. We define each instance of raw EEG timeseries as $\textbf{X}_{\text{raw}} \in \mathbb{R}^{ C \times T}$ with $C$ channels and time step of $T$. The EEG data were first segmented into non-overlapping epochs by calculating the epoch length $T^{\prime}$ in samples and determining the total number of epochs that could be extracted. For each epoch, the algorithm extracted temporal segments across all channels and performed channel-wise spectral analysis. We compute the power spectral density (PSD) using Welch's method and define the total power over the 0.5–45 Hz range.
We then compute relative band power (RBP) in four non-overlapping bands $\delta$ (0.5–4 Hz), $\theta$ (4–8 Hz), $\alpha$ (8–13 Hz), and $\beta$ (13–45 Hz),
using left-closed, right-open intervals so that bands partition the total power. 

Given the extracted feature set $\{(\mathbf{X}_i, y_i)\}_{i=1}^{N}$, where $\mathbf{X}_i \in \mathbb{R}^{K \times C \times F}$ denotes the RBP feature tensor for the $i$-th sample, composed of $K = 4$ frequency-band feature matrices $\mathbf{X}_i^{(k)} \in \mathbb{R}^{C \times F}$ across $C = 19$ channels, with $F$ being the length of the RBP-derived feature sequence, and $y_i \in \{0, 1\}$ is the corresponding binary class label.

\vspace{-2mm}
\subsection{Multi-Granularity Transformer for Node Feature Extraction}

Aiming to capture multi-scale temporal patterns in RBP-derived EEG features while extracting meaningful node representations for each channel, the MGFormer ~\cite{rahman2025mgformer} architecture was designed to more effectively capture multi-level features encompassing
both local and long-range temporal signals. As shown in Fig.~\ref{fig:Transformer} of the MGT-NFE components, these include self-attention to capture long-range temporal dependencies, a one-dimensional convolutional neural
network (1-D CNN) to extract local temporal details, and FFT features.

\subsubsection{Multi-Granularity Token Encoder}

For each channel--band pair $(c, k)$ with $k \in \{\delta, \theta, \alpha, \beta\}$, the corresponding RBP feature sequence $\mathbf{x}_{c,k} \in \mathbb{R}^{F}$ is defined as the $c$-th row vector of $\mathbf{X}_i^{(k)}$. Multi-scale temporal patterns are then extracted from this RBP feature sequence using convolutional layers with kernel sizes tailored to different temporal granularities. Let \(\mathcal{G}\) denote the set of temporal granularity indices, and each granularity \(g \in \mathcal{G}\) is associated with a one-dimensional convolutional kernel of length \(\Phi(g)\), a stride \(s(g)\), and a pooling operator \(P_g(\cdot)\). Given the RBP feature sequence \(\mathbf{x}_{c,k} \in \mathbb{R}^{F}\), the temporal token at granularity \(g\) is defined as
\begin{equation}
\mathbf{T}^{(c,k)}_{g}
=
P_g\!\Big(
\sigma\!\big(
\mathrm{Conv1D}
(
\mathbf{x}_{c,k};\,\Phi(g),\,s(g)
)
\big)
\Big)
\in
\mathbb{R}^{L_g\times D_T},
\end{equation}
where $\sigma(\cdot)$ is a nonlinear activation function, such as LeakyReLU, 
$D_T$ is the number of convolutional filters, and $L_g$ is the resulting token length after convolution, stride, and pooling. 
By concatenating the tokens across all granularities, we obtain
\begin{equation}
\mathbf{T}^{(c,k)}
=
\mathrm{Concat}_{g\in\mathcal{G}}
\left[
\mathbf{T}^{(c,k)}_{g}
\right]
\in
\mathbb{R}^{L\times D_T},
\qquad
L=\sum_{g\in\mathcal{G}} L_g .
\end{equation}

\subsubsection{Transformer Encoding}

To preserve temporal ordering, we add a positional encoding \(\mathbf{PE}\in\mathbb{R}^{L\times D_T}\) to the concatenated tokens before feeding them into the Transformer layers:
\begin{align}
\mathbf{E}^{(0)} &= \mathbf{T}^{(c,k)} + \mathbf{PE},\\
\mathbf{E}^{(l+1)} &= \mathrm{TransformerBlock}\left(\mathbf{E}^{(l)}\right), \quad l=0,1,\ldots,L_{\text{tr}}-1,
\end{align}
where \(L_{\text{tr}}\) denotes the total number of Transformer layers. Each \(\mathrm{TransformerBlock}(\cdot)\) consists of multi-head self-attention and position-wise feed-forward networks, producing outputs of the same shape \(\mathbb{R}^{L\times D_T}\).

After $L_{\mathrm{tr}}$ transformer layers, we obtain the final encoded features $\mathbf{E}^{(L_{\mathrm{tr}},c,k)}$ for each channel--band pair. We then extract a representative per-channel, per-band node feature from the final transformer output:
\begin{equation}
\mathbf{h}_{c}^{(k)} = \operatorname{Aggregate}\!\left(\mathbf{E}^{(L_{\mathrm{tr}},c,k)}\right) \in \mathbb{R}^{d_h},
\end{equation}
where $\operatorname{Aggregate}(\cdot)$ can be mean pooling, max pooling, or learned attention pooling, and $d_h$ denotes the per-band node feature dimension.

Applying the MGT-NFE independently to each frequency band and stacking the resulting per-channel features yields the \emph{band-specific} node feature matrix:
\begin{equation}\label{eq:per_band_H}
\mathbf{H}^{(k)} = [\mathbf{h}_{1}^{(k)}; \mathbf{h}_{2}^{(k)}; \ldots; \mathbf{h}_{C}^{(k)}] = \operatorname{MGT\text{-}NFE}(\mathbf{X}^{(k)}_{i}),
\end{equation}
where $\mathbf{H}^{(k)} \in \mathbb{R}^{C \times d_{h}}$ represents the node feature matrix for frequency band $k$. To further obtain a unified cross-band descriptor, we concatenate the band-wise features of each channel:
\begin{equation}\label{eq:concat_h}
\mathbf{h}_{c} = \operatorname{Concat}[\mathbf{h}_{c}^{(\delta)}, \mathbf{h}_{c}^{(\theta)}, \mathbf{h}_{c}^{(\alpha)}, \mathbf{h}_{c}^{(\beta)}],
\end{equation}
where $\mathbf{h}_c  \in \mathbb{R}^{D^{\prime}} $ with $D^{\prime}=4d_h$ being the total node feature dimension. The stacking over channels gives the unified node feature matrix $\mathbf{H} = [\mathbf{h}_{1}; \mathbf{h}_{2}; \ldots; \mathbf{h}_{C}] \in \mathbb{R}^{C \times D'}$. 




\subsection{GMRF Prior Modeling}
In this section, we formulate the extracted node features $\mathbf{H}^{(k)}$ for each band $k$ as a corresponding band graph $G^{(k)}=\{(\mathcal{E}^{(k)},\mathcal{V}^{(k)}), \mathbf{A}^{(k)}\}$ with vertex $\mathcal{V}^{(k)} = \{v_{i}\}_{i=1}^{C}$ and adjacency matrix $\mathbf{A}^{(k)} \in \mathbb{R}^{C \times C}$. As shown in Fig.~\ref{fig:Fig1}, our goal is to establish a gating network from a variational perspective as the most suitable MoE weighting mechanism. Existing EEG research has demonstrated that the characteristic differences across different frequency bands in early AD can serve as clinical biomarkers for diagnostic interpretation~\cite{jiao2023neural,meghdadi2021resting}. To characterize and encode different band graph features, we introduce latent variables $\mathbf{Z}^{(k)} \in \mathbb{R}^{C \times d_{z}}$ for each band $k$, where $d_z$ is the latent dimension. We denote by $\mathbf{z}^{(k)}_{d} \in \mathbb{R}^{C}$ the $d$-th column of $\mathbf{Z}^{(k)}$, which collects the values of all $C$ nodes along the $d$-th latent dimension ($d = 1, \dots, d_z$). We model each latent band graph as having a prior distribution that follows a  GMRF. The precision matrix $\mathbf{Q}^{(k)} \in \mathbb{R}^{C \times C}$ encodes the graph structure and satisfies the Markov property, has edges $(i, j)^{(k)} \in \mathcal{E}^{(k)} \Longleftrightarrow Q_{ij}^{(k)} \neq 0$ for all $i \neq j$. In practice, the precision matrix $\mathbf{Q}^{(k)}$ is derived from the symmetric normalized graph Laplacian $\mathbf{L}$ with an added diagonal loading term:

\begin{equation}\label{eq:precision_matrix}
\mathbf{Q}^{(k)} = \mathbf{L}^{(k)} + \lambda_{Q}\mathbf{I}, 
\qquad 
\mathbf{L}^{(k)} = \mathbf{I} - \mathbf{D}^{(k)^{-1/2}}\mathbf{A}^{(k)}\mathbf{D}^{(k)^{-1/2}},
\end{equation}
where $\mathbf{I} \in \mathbb{R}^{C \times C}$ is the identity matrix and positive semi-definite, $\mathbf{D}^{(k)} \in \mathbb{R}^{C \times C}$ is the degree matrix of $\mathbf{A}^{(k)}$, and $\lambda_{Q} > 0$ is a small diagonal loading constant. The term $\lambda_{Q}\mathbf{I}$ ensures that $\mathbf{Q}^{(k)}$ is positive definite, so that its inverse $(\mathbf{Q}^{(k)})^{-1}$ exists and $\log|\mathbf{Q}^{(k)}|$ is well defined. The structured GMRF prior is then applied independently to each latent dimension:

\begin{equation}\label{eq:gmrf_prior}
p(\mathbf{Z}^{(k)} \mid \mathbf{A}^{(k)}) = \prod_{d=1}^{d_z}\mathcal{N}(\mathbf{z}^{(k)}_{d}; \mathbf{0}, (\mathbf{Q}^{(k)})^{-1}),
\end{equation}
in which the prior distribution encodes the smoothness assumption of the graph structure, where two nodes strongly connected in $\mathbf{A}^{(k)}$ are encouraged to have similar latent representations in $\mathbf{Z}^{(k)}$, enabling the model to leverage EEG brain connectivity patterns specific to each frequency band.

\subsection{Variational Graph Convolutional Encoder}

In this work, we design a simple two-layer graph convolutional network (GCN) to encode the node features $\mathbf{H}^{(k)}$ from the MGT-NFE of the \(k\)-th frequency band into latent representations. Specifically, given the adjacency matrix \(\mathbf{A}^{(k)}\) and node feature matrix \(\mathbf{H}^{(k)}\), the output of the GCN encoder is computed as:

\begin{equation}
    \tilde{\mathbf{H}}^{(k)}= \text{GCN}^{(k)}(\mathbf{H}^{(k)}, \mathbf{\hat{A}}^{(k)}) 
\end{equation}
where \(\hat{\mathbf{A}}^{(k)} = \mathbf{D}^{(k)^{-1/2}} \mathbf{A}^{(k)} \mathbf{D}^{(k)^{-1/2}}
\) is the symmetrically normalized adjacency matrix. These GCN embeddings $\tilde{\mathbf{H}}^{(k)}=[\tilde{\mathbf{h}}^{(k)}_{1},\tilde{\mathbf{h}}^{(k)}_{2},...,\tilde{\mathbf{h}}^{(k)}_{C}]$ serve as inputs to a node-level variational encoder, which approximates the posterior distribution over latent variables \( \mathbf{Z}^{(k)} = [\mathbf{z}_1^{(k)}, \dots, \mathbf{z}_C^{(k)}]^{\top} \), where $\mathbf{z}_i^{(k)} \in \mathbb{R}^{d_z}$ denotes the latent vector of node $i$. The graph encoder produces a mean-field Gaussian variational posterior that is factorized over nodes:

\begin{equation}\label{eq:posterior_node}
q_{\phi}(\mathbf{Z}^{(k)} \mid \mathbf{H}^{(k)}, \mathbf{A}^{(k)}) = 
\prod_{i=1}^{C} \mathcal{N} \left( 
\mathbf{z}_i^{(k)} \mid \boldsymbol{\mu}_i^{(k)}, 
\operatorname{diag} \left( (\boldsymbol{\sigma}_i^{(k)})^2 \right) 
\right),
\end{equation}
where the mean \( \boldsymbol{\mu}_i^{(k)} \in \mathbb{R}^{d_z} \) and log standard deviation \( \log \boldsymbol{\sigma}_i^{(k)} \in \mathbb{R}^{d_z} \) are computed from the node embedding $\tilde{\textbf{h}}^{(k)}_{i}$ of the $k$-th expert via simple multi-layer perceptron (MLP) mappings:

\begin{equation}
\boldsymbol{\mu}_i^{(k)} = \mathbf{W}_\mu \, \tilde{\mathbf{h}}_i^{(k)}, \quad  
\log \boldsymbol{\sigma}_i^{(k)} = \mathbf{W}_{\log\sigma} \, \tilde{\mathbf{h}}_i^{(k)},
\end{equation}
where $\mathbf{W}_{\mu}, \mathbf{W}_{\log\sigma} \in \mathbb{R}^{d_z \times H_{\text{gcn}}}$ project 
the GCN output $\tilde{\mathbf{h}}_{i}^{(k)} \in \mathbb{R}^{H_{\text{gcn}}}$ to the latent dimension $d_z$. 

To enable gradient-based optimization through stochastic sampling, we apply the \textit{reparameterization trick}~\cite{kingma2014auto} to obtain:

\begin{equation}
\mathbf{z}_i^{(k)} = \boldsymbol{\mu}_i^{(k)} + \boldsymbol{\sigma}_i^{(k)} \odot \boldsymbol{\epsilon}_i, \quad \boldsymbol{\epsilon}_i \sim \mathcal{N}(\mathbf{0}, \mathbf{I}),
\end{equation}
where \( \odot \) denotes element-wise multiplication. This guarantees the sampling process to be differentiable and supports efficient backpropagation during training. The resulting latent representations \( \mathbf{z}_i^{(k)} \) are then used for downstream tasks, such as graph-level classification through mean pooling. 


\section{Variational Mixture of Graph-Structured Experts}\label{sec:VMoGE}

\subsubsection{Graph-Level Representation and Expert Decoder}

For classification, we aggregate the node-level latent representations using mean pooling:
\begin{equation}
\bar{\mathbf{z}}^{(k)} = \frac{1}{C} \sum_{i=1}^{C} \mathbf{z}_i^{(k)} \in \mathbb{R}^{d_z}
\end{equation}
where $\bar{\mathbf{z}}^{(k)}$ is then passed through an expert-specific decoder network to produce the $k$-th expert's logit output:
\begin{equation}
\hat{y}^{(k)} = \text{MLP}(\bar{\mathbf{z}}^{(k)}; \boldsymbol{\theta}_k)
\end{equation}
where $\boldsymbol{\theta}_k$ represents the learnable parameters of the $k$-th expert decoder.

\subsubsection{Mixture of Experts with Gating Network}

In contrast to conventional fixed mixture models~\cite{jacobs1991adaptive, graphdive}, we propose an adaptive gating mechanism grounded in graph structural priors within the MoE framework, which enables each input sample to dynamically modulate the relative contribution of individual frequency-band experts based on the underlying graph-prior information. Formally, the gating coefficients for the $k$-th expert are obtained through a softmax operation applied over all experts, and can be formulated as:

\begin{equation}
\pi^{(k)}(\mathbf H')=
\frac{
\exp\left(
\mathbf w_k^{\top}
f_{\boldsymbol{\psi}}(\mathbf H')
\right)
}{
\sum_{j=1}^{K}
\exp\left(
\mathbf w_j^{\top}
f_{\boldsymbol{\psi}}(\mathbf H')
\right)
}.
\end{equation}
where $f_{\boldsymbol{\psi}}(\mathbf H')$ denotes a gating feature mapping applied to the concatenated input $\mathbf{H}^{\prime} = \text{Concat}([\mathbf{H}^{(1)}, \mathbf{H}^{(2)}, \ldots, \mathbf{H}^{(K)}])$, and $\mathbf{w}_k$ is the parameter vector associated with the $k$-th expert. Then, the final prediction is computed as a weighted combination of all expert outputs:

\begin{equation}\label{eq:mixture_output}
\hat{y} = \sum_{k=1}^{K} \pi^{(k)}(\mathbf{H}^{\prime}) \cdot \hat{y}^{(k)}
\end{equation}
where the softmax ensures $\sum_{k=1}^{K} \pi^{(k)}(\mathbf{H}^{\prime})= 1$  and $\pi^{(k)}(\mathbf{H}^{\prime}) \in [0, 1]$. 

\subsection{Predictive Distribution for VMoGE}

Our predictive distribution over the output label $y$ integrates over the latent variables from all frequency band experts, weighted by the input-dependent gating mechanism. Formally, the predictive distribution can be expressed as:

\begin{align}\label{eq:mixture}
p_{\theta}(y \mid \{ \mathbf{H}^{(k)},\mathbf{A}^{(k)}\}_{k=1}^{K}; \boldsymbol{\Theta}) 
&= \sum_{k=1}^{K} \pi^{(k)}(\mathbf{H}^{\prime}) \notag \\
&\quad \hspace{-7em} \times \mathbb{E}_{q_{\boldsymbol{\phi}}(\mathbf{Z}^{(k)} \mid \mathbf{H}^{(k)}, \mathbf{A}^{(k)})} \left[ p(y \mid \mathbf{Z}^{(k)}; \boldsymbol{\theta}_k) \right],
\end{align}
where $\boldsymbol{\Theta} = \{\{\boldsymbol{\theta}_k\}_{k=1}^{K}, \boldsymbol{\phi}\}$ denotes the complete set of model parameters, including the expert-specific parameters $\boldsymbol{\theta}_k$, and encoder parameters $\boldsymbol{\phi}$.
Here, gating weights $\pi^{(k)}(\mathbf{H}^{\prime})$ are computed dynamically for each input.  Here, the likelihood term $p(y \mid \mathbf{Z}^{(k)}; \boldsymbol{\theta}_k)$ represents the conditional probability of the class label given the latent representation for the $k$-th expert.

\subsection{Optimizing Variational Structured Inference Lower Bounds}

Unlike traditional optimization methods that use lower bounds, we incorporate the structured properties of each band's relational graph into the variational lower bound objective. We aim to effectively approximate the latent posterior distributions and preserve the structural dependencies embedded in the graph data.
Specifically, we formulate our training objective based on the ELBO, which balances two critical components: predictive performance and latent regularization. Our goal is to maximize the ELBO defined as:

\begin{align}\label{eq:ELBO}
\max_{\boldsymbol{\Theta}} \ \mathcal{L}_{\mathrm{ELBO}} 
=&\;   
\mathbb{E}_{q_{\boldsymbol{\phi}}}\left[\log
p_{\boldsymbol{\theta}}
\left(
y
\mid
\{\mathbf{Z}^{(k)}\}_{k=1}^{K},
\{\mathbf{H}^{(k)}\}_{k=1}^{K}; \boldsymbol{\Theta}
\right)\right] \notag  \\
& \hspace{-5em}  - \lambda \cdot \sum_{k=1}^{K}
D_{\mathrm{KL}}^{(k)}\left(q_{\phi}(\mathbf{Z}^{(k)} \mid \mathbf{H}^{(k)},\mathbf{A}^{(k)}) \ \| \ p(\mathbf{Z}^{(k)} \mid \mathbf{A}^{(k)})\right ) \notag \\
\end{align}
where $\lambda$ is a regularization parameter, and the first term maximizes the expected log-likelihood of the target label under the joint variational posterior distribution over the band-specific latent representations. We assume that this posterior factorizes across the
\(K\) frequency bands as
$q_{\boldsymbol{\phi}}
\left(
\{\mathbf{Z}^{(k)}\}_{k=1}^{K}
\mid
\{\mathbf{H}^{(k)},\mathbf{A}^{(k)}\}_{k=1}^{K}
\right)
=
\prod_{k=1}^{K}
q_{\boldsymbol{\phi}}^{(k)}
\left(
\mathbf{Z}^{(k)}
\mid
\mathbf{H}^{(k)},\mathbf{A}^{(k)}
\right)$. 
The second term imposes a KL divergence that regularizes 
the learned latent representations toward the conditional mixture of GMRF 
priors $p(\mathbf{Z}^{(k)} \mid \mathbf{A}^{(k)})$ defined in Eq.~\ref{eq:gmrf_prior}, 
ensuring that the latent space respects the structural dependencies encoded 
in each band-specific graph. The coefficient $\lambda$ controls the strength 
of this structural regularization.

\paragraph{\textbf{GMRF Expert KL Divergence}}
To model structured prior information from band-specific relational graphs, we construct a structured variational posterior distribution $q_{\phi}(\mathbf{Z}^{(k)} \mid \mathbf{H}^{(k)}, \mathbf{A}^{(k)})$ and a prior distribution $p(\mathbf{Z}^{(k)} \mid \mathbf{A}^{(k)})$ to formulate the KL divergence $ D_{KL}(q(\cdot) \mid\mid  p(\cdot))$. As established in Eq.~\eqref{eq:posterior_node} and Eq.~\eqref{eq:gmrf_prior}, the posterior factorizes over nodes while the GMRF prior factorizes over the $d_z$ latent dimensions. Both formulations correspond to a matrix-variate Gaussian over $\mathbf{Z}^{(k)}$ with independent columns. We therefore evaluate the KL divergence per latent dimension and sum over the $d_z$ dimensions. Specifically, for the $d$-th latent dimension, the variational posterior over the $C$-node column $\mathbf{z}_d^{(k)} \in \mathbb{R}^{C}$ admits the marginal variational posterior
\begin{equation}\label{eq:kl_posterior_dim}
    q_{\phi}(\mathbf{z}_{d}^{(k)}) = \mathcal{N}\!\left(\mathbf{z}_{d}^{(k)}; \boldsymbol{\mu}_{d}^{(k)}, \boldsymbol{\Sigma}_{d}^{(k)}\right),
\end{equation}
where covariance $\boldsymbol{\Sigma}_{d}^{(k)} = \operatorname{diag}\!\left([(\sigma_{1,d}^{(k)})^2, \ldots, (\sigma_{C,d}^{(k)})^2]\right) \in \mathbb{R}^{C \times C}$ is diagonal by the mean-field assumption, with its $i$-th entry $(\sigma_{i,d}^{(k)})^2$ denoting the marginal variance of node $i$ along the $d$-th latent dimension. The expert-specific GMRF prior incorporates the structural information from the adjacency matrix $\mathbf{A}^{(k)}$ through the precision matrix $\mathbf{Q}^{(k)}$:
\begin{equation}\label{eq:kl_prior_dim}
    p(\mathbf{z}_{d}^{(k)}) = \mathcal{N}(\mathbf{z}_{d}^{(k)}; \mathbf{0}, (\mathbf{Q}^{(k)})^{-1}),
\end{equation}
where $\mathbf{Q}^{(k)}  \in \mathbb{R}^{C \times C}$ is the positive-definite precision matrix from Eq.~\eqref{eq:precision_matrix}. The KL divergence between $q_{\phi}(\mathbf{z}_d^{(k)})$ and $p(\mathbf{z}_d^{(k)})$ for a single dimension is given by

\begin{align}
D_{KL}^{(d)} 
&= \mathbb{E}_{q_{\phi}(\mathbf{z}_d^{(k)})} \left[ \log \frac{q_{\phi}(\mathbf{z}_d^{(k)})}{p(\mathbf{z}_d^{(k)})} \right] \nonumber \\
&= \mathbb{E}_{q_{\phi}(\mathbf{z}_d^{(k)})} \left[ -\frac{1}{2}\log|\boldsymbol{\Sigma}_d^{(k)}| \right. \nonumber \\
&\quad \left. - \frac{1}{2}(\mathbf{z}_d^{(k)} - \boldsymbol{\mu}_d^{(k)})^{\top} (\boldsymbol{\Sigma}_d^{(k)})^{-1} (\mathbf{z}_d^{(k)} - \boldsymbol{\mu}_d^{(k)}) \right. \nonumber \\
&\quad \left. + \frac{1}{2}\log|(\mathbf{Q}^{(k)})^{-1}| + \frac{1}{2}(\mathbf{z}_d^{(k)})^{\top} \mathbf{Q}^{(k)} \mathbf{z}_d^{(k)} \right],
\end{align}
More specifically, we compute the expectations of the quadratic forms. For the centered quadratic form under $q_{\phi}(\mathbf{z}_d^{(k)})$, since $\mathbf{z}_d^{(k)} \sim \mathcal{N}(\boldsymbol{\mu}_d^{(k)}, \boldsymbol{\Sigma}_d^{(k)})$:

\begin{align}\label{eq:first_quadratic_term}
&\mathbb{E}_{q_{\phi}(\mathbf{z}_d^{(k)})}
\left[(\mathbf{z}_d^{(k)} - \boldsymbol{\mu}_d^{(k)})^{\top} (\boldsymbol{\Sigma}_d^{(k)})^{-1} (\mathbf{z}_d^{(k)} - \boldsymbol{\mu}_d^{(k)})\right] \nonumber \\
&\qquad = \mathbb{E}_{q_{\phi}(\mathbf{z}_d^{(k)})}\!\left[\text{tr}\!\left((\boldsymbol{\Sigma}_d^{(k)})^{-1} (\mathbf{z}_d^{(k)} - \boldsymbol{\mu}_d^{(k)})(\mathbf{z}_d^{(k)} - \boldsymbol{\mu}_d^{(k)})^{\top}\right)\right] \nonumber \\
&\qquad = \text{tr}(\mathbf{I}_{C}) = C.
\end{align}

For the non-centered quadratic form involving the prior precision, we obtain
\begin{align}\label{eq:quad_form_2}
\mathbb{E}_{q_{\phi}(\mathbf{z}_d^{(k)})}[(\mathbf{z}_d^{(k)})^{\top} \mathbf{Q}^{(k)} \mathbf{z}_d^{(k)}] 
&= \mathbb{E}_{q_{\phi}(\mathbf{z}_d^{(k)})}[\text{tr}(\mathbf{Q}^{(k)} \mathbf{z}_d^{(k)}(\mathbf{z}_d^{(k)})^{\top})] \nonumber \\
&= \text{tr}(\mathbf{Q}^{(k)} \boldsymbol{\Sigma}_d^{(k)}) + (\boldsymbol{\mu}_d^{(k)})^{\top} \mathbf{Q}^{(k)} \boldsymbol{\mu}_d^{(k)}. 
\end{align}
Combining Eq.~\eqref{eq:first_quadratic_term} and Eq.~\eqref{eq:quad_form_2}, and summing over all $d_z$ latent dimensions, we obtain the closed-form expression for the $k$th expert-specific KL divergence:

\begin{align}
D^{(k)}_{\mathrm{KL}}
&= \sum_{d=1}^{d_z} D_{KL}^{(k,d)} \nonumber \\
&= \frac{1}{2} \sum_{d=1}^{d_z} \bigg[ 
\mathrm{tr} \!\left( \mathbf{Q}^{(k)} \boldsymbol{\Sigma}_d^{(k)} \right)
+ (\boldsymbol{\mu}_d^{(k)})^{\top} \mathbf{Q}^{(k)} \boldsymbol{\mu}_d^{(k)} 
- C \nonumber \\
&\quad - \log \left| \boldsymbol{\Sigma}_d^{(k)}\mathbf{Q}^{(k)} \right| 
\bigg],
\label{eq:final_kl}
\end{align}
where the trace term $\mathrm{tr}(\mathbf{Q}^{(k)} \boldsymbol{\Sigma}_d^{(k)})$ penalizes posterior uncertainty according to the graph-structured prior precision, the quadratic term $(\boldsymbol{\mu}_d^{(k)})^{\top} \mathbf{Q}^{(k)} \boldsymbol{\mu}_d^{(k)}$ captures the deviation of the posterior mean from the prior mean under the graph-structured precision, and the log-determinant of the diagonal posterior covariance is $\log|\boldsymbol{\Sigma}_d^{(k)}| = \sum_{i=1}^{C} \log (\sigma_{i,d}^{(k)})^2$. Because $\mathbf{Q}^{(k)}$ is shared across the $d_z$ dimensions, the term $\log|\mathbf{Q}^{(k)}|$ is computed once and reused; in practice we evaluate it via a Cholesky factorization $\mathbf{Q}^{(k)} = \mathbf{L}\mathbf{L}^{\top}$ as $\log|\mathbf{Q}^{(k)}| = 2\sum_{i} \log L_{ii}$ for numerical stability.

\paragraph{\textbf{Overall Optimization}}
In practice, we implement the negative ELBO as our overall loss function for the training stage as follows:
\begin{equation}\label{eq:total_loss}
\mathcal{L}_{\text{total}} = \mathcal{L}_{\text{cls}} + \lambda \cdot \mathcal{L}_{\text{KL}}.
\end{equation}
where $\mathcal{L}_{\text{cls}}$ denotes the binary classification loss computed from the weighted mixture of expert outputs in Eq.~\eqref{eq:mixture_output}, and $\mathcal{L}_{\text{KL}} = \sum_{k=1}^{K} D_{\text{KL}}^{(k)}$ aggregates the expert-specific KL divergences in Eq.~\eqref{eq:final_kl} over all $K$ experts. The hyperparameter $\lambda$ controls the strength of the structural regularization.

\section{Experiments}\label{sec:experiments}
To evaluate our proposed VMoGE on two EEG datasets, we designed our experiments to address our research questions as follows:



\begin{itemize}
    \item \textbf{RQ1:} 
    Does the proposed VMoGE framework improve AUC and accuracy for EEG-based AD and FTD diagnosis compared to state-of-the-art feature extraction and graph-based models?

    \item \textbf{RQ2:} 
    How does integrating multiple frequency-band experts within the MoE framework influence the discriminative power and robustness of EEG-based dementia classification?

    \item \textbf{RQ3:}
    Can the incorporation of GMRF priors enhance model stability and generalization across heterogeneous datasets, especially under limited-sample conditions?

    \item \textbf{RQ4:}
    Do the expert gating weights learned by VMoGE reflect clinically meaningful associations with cognitive indicators (e.g., MMSE scores) and age-related EEG spectral changes, thus providing interpretable biomarkers for dementia progression?
\end{itemize}

\subsection{Data Acquisition}

\begin{itemize}

    \item \textbf{Open AD dataset~\cite{miltiadous2023dataset}:} The EEG dataset contained resting-state recordings from 88 elderly subjects: 36 AD patients, 23 FTD patients, and 29 healthy controls (HC). EEG signals were acquired using a 19-channel clinical system (10-20 electrode placement) at 500 Hz during eyes-closed conditions, with recording durations averaging 12-14 minutes per subject. Participants had mean MMSE scores of 17.75 (AD), 22.17 (FTD), and 30 (HC). The dataset includes both raw and preprocessed data with bandpass filtering (0.5-45 Hz), artifact subspace reconstruction, and independent component analysis for denoising.
    
    \item \textbf{Session-based AD~\cite {yang2013cognitive}:} To validate the proposed method, this study included 123 participants recruited from Taipei Veterans General Hospital, excluding those with secondary causes of dementia or a history of major psychiatric disorders. Patients were categorized into three CDR types: CDR=0 ($n=15$), CDR=1 ($n=84$), and CDR$\geq$2 ($n=24$). The AD patients were diagnosed with probable Alzheimer's disease according to the  National Institute of Neurological and Communicative Disorders and the Stroke/Alzheimer's Disease and Related Disorders Association diagnostic criteria~\cite{mckhann1984clinical}, with a mean age of 78.0 years (SD=8.6). EEG data were recorded using the Nicolet EEG system. Participants underwent a five-minute adaptation period before the test, followed by three alternating 10- to 20-second closed-eye and open-eye segments, along with a light stimulation test. Electrodes were placed according to the international 10--20 system, totaling 19 leads. The sampling frequency was 256 Hz, with high-pass and low-pass filters set to 0.05 Hz and 70 Hz, respectively, and a 60 Hz notch filter. Electrode impedance was maintained below 3 k$\Omega$. All signals used the earlobe as the reference point. Technicians monitored subjects' alertness throughout testing, prompting them to stay awake when signs of drowsiness appeared. After manual review, three 10-second segments free of significant noise were extracted from the closed-eye phase of the raw EEG data for subsequent data collection.

\end{itemize}

\subsection{EEG Data Preprocessing}

The raw EEG signals underwent a systematic multi-stage preprocessing pipeline to ensure signal quality and remove artifacts. Initially, band-pass filtering between 0.05 and 70 Hz was applied using a finite impulse response   filter with a Hamming window, followed by a 60 Hz notch filter to eliminate power-line interference. The filtered signals were re-referenced to averaged bilateral mastoid electrodes (A1/A2), with average referencing employed when mastoid electrodes were unavailable. Artifact removal was conducted in two stages: first, Artifact Subspace Reconstruction  with a 0.5-second sliding window and 17 standard deviation cutoff threshold was applied to correct high-amplitude transient artifacts; second, independent component analysis using the Infomax algorithm was performed, with ICLabel-classified ocular and myogenic components automatically excluded from signal reconstruction. Following artifact removal, the cleaned continuous EEG data were segmented into consecutive, non-overlapping 6-second epochs, with incomplete terminal segments discarded. Signals were decomposed into frequency bands under two configurations: (1) four canonical bands: $\delta$ (0.5--4 Hz), $\theta$ (4--8 Hz), $\alpha$ (8--13 Hz), and $\beta$ (13--45 Hz); and (2) an extended five-band configuration in which the $\beta$ band is split into $\beta_1$ (13--25 Hz) and $\gamma$ (25--45 Hz), as used in the experiments reported in Table.~\ref{table:overall}.

The PSD was estimated for each epoch and channel using Welch's method with 4-second FFT windows spanning 0.5 to 45 Hz. RBP features were computed by normalizing band-specific power by total power within 0.5-45 Hz. Extracted features were organized into subject-level tensors with zero-padding applied to equalize epoch counts across participants, and age and MMSE scores were aligned with EEG features for multimodal analyses. 

\begin{table}
\centering
\caption{Hyper-parameter settings of VMoGE.}
\renewcommand{\arraystretch}{1.05}
\setlength{\tabcolsep}{3pt}
\begin{tabular}{p{2.8cm} p{5.2cm}}
\hline
\textbf{Parameter} & \textbf{Values} \\
\hline
EEG channels & $C = 19$ \\
Epoch length & $T = 6\,\mathrm{s}$ \\
Kernel ratios & $\mathcal{G}$ =$\{0.04, 0.06, 0.08\}\times SR^{\dagger}$ \\
CNN kernel size & $k_{\mathrm{cnn}} = 11$ \\
Temporal filters & $D_T = 32$ \\
Pooling stride & $s_p = 4$ \\
Embedding dim. & $d = 64$ \\
Transformer depth & $L_{\mathrm{tr}} = 2$ \\
Attention heads & 8 \\
GCN hidden dim. & $H_{\mathrm{gcn}} = 64$ \\
Node feature dim. & $D' = 768$ per channel \\
Training stage & Adam $(\eta = 10^{-3})$, $E = 200$ epochs \\
\hline
\end{tabular}
\label{tab:params}
\begin{flushleft}
\footnotesize
$^{\dagger}$ denotes the sampling rate (SR) of the EEG signal.
\end{flushleft}
\label{tab:parameters_setting}
\vspace{-3mm}
\end{table}

\begin{table*}
\centering
\huge
\caption{Comparative analysis of VMoGE for six EEG dementia subtype classifications. Performance metrics, including Area Under Curve (AUC) and Accuracy (ACC), are evaluated in the Open AD and Session-based settings across three classification tasks. The best results are highlighted in \textbf{bold}, and the second-best results are \underline{underlined}.}
\label{tab:trinetx_results}
\begin{threeparttable}
\renewcommand{\arraystretch}{1.15}
\begin{adjustbox}{max width=\textwidth}
\begin{tabular}{lrrrrrrrrrrrr}
\toprule
\multicolumn{1}{l}{\textbf{Model}} &
\multicolumn{4}{c}{\textbf{HC vs FTD / CDR=0 vs CDR=1}} &
\multicolumn{4}{c}{\textbf{HC vs AD / CDR=0 vs CDR=2}} &
\multicolumn{4}{c}{\textbf{FTD vs AD / CDR=1 vs CDR=2}} \\
\cmidrule(lr){2-5} \cmidrule(lr){6-9} \cmidrule(lr){10-13}
& \multicolumn{2}{c}{\textit{Open AD}} & \multicolumn{2}{c}{\textit{Session-based AD} }
& \multicolumn{2}{c}{\textit{Open AD}} & \multicolumn{2}{c}{\textit{Session-based} AD}
& \multicolumn{2}{c}{\textit{Open AD}} & \multicolumn{2}{c}{\textit{Session-based} AD } \\
\cmidrule(lr){2-3}\cmidrule(lr){4-5}\cmidrule(lr){6-7}\cmidrule(lr){8-9}\cmidrule(lr){10-11}\cmidrule(lr){12-13}
& \multicolumn{1}{c}{AUC} & \multicolumn{1}{c}{ACC} & \multicolumn{1}{c}{AUC} & \multicolumn{1}{c}{ACC}
& \multicolumn{1}{c}{AUC} & \multicolumn{1}{c}{ACC} & \multicolumn{1}{c}{AUC} & \multicolumn{1}{c}{ACC}
& \multicolumn{1}{c}{AUC} & \multicolumn{1}{c}{ACC} & \multicolumn{1}{c}{AUC} & \multicolumn{1}{c}{ACC} \\
\midrule
EEGNet~\cite{lawhern2018eegnet}     & 0.64 ± 0.25 & 0.62 ± 0.17 & 0.53 ± 0.04 & 0.49 ± 0.09  & 0.81 ± 0.11 & 0.76 ± 0.13 & 0.52 ± 0.15 & 0.54 ± 0.12  &  0.71 ± 0.19 & 0.69 ± 0.13 & \underline{0.59 ± 0.04} & 0.57 ± 0.02  \\
EEGViT~\cite{yang2023vit2eeg}      & 0.64 ± 0.20 & 0.56 ± 0.15 & 0.45 ± 0.10 & 0.41 ± 0.08  & 0.83 ± 0.14 & 0.76 ± 0.17 &  0.60 ± 0.14 & 0.58 ± 0.04  & 0.68 ± 0.18 & 0.61 ± 0.11 & 0.58 ± 0.12  & 0.53 ± 0.08  \\
DTCA-Net~\cite{dharia2026dual}         & 0.55 ± 0.08 & 0.63 ± 0.10 &  0.46 ± 0.15 & \underline{0.56 ± 0.12} & 0.83 ± 0.10 &  0.74 ± 0.08 &  0.62 ± 0.09 &  0.62 ± 0.07 & 0.62 ± 0.23 & 0.58 ± 0.19 & 0.50 ± 0.06 &  \underline{0.62 ± 0.07} \\
Deformer~\cite{ding2024eeg}       & \underline{0.75 ± 0.14} & \underline{0.72 ± 0.12} & 0.41 ± 0.13 & 0.49 ± 0.06  & 0.84 ± 0.18 & \underline{0.80 ± 0.13} & 0.59 ± 0.16 & 0.61 ± 0.14  & 0.68 ± 0.20 & 0.61 ± 0.15 & 0.59 ± 0.10 & \underline{0.62 ± 0.07}  \\
ADFormer~\cite{wang2024ADFormer}       & 0.70 ± 0.21 & 0.61 ± 0.17 & 0.47 ± 0.06  & 0.53 ± 0.10  & 0.85 ± 0.23 & 0.78 ± 0.17 & 0.45 ± 0.15 & 0.44 ± 0.08  & \underline{0.77 ± 0.14} & \textbf{0.74 ± 0.15}  & 0.43 ± 0.10 & 0.41 ± 0.11  \\
MGformer~\cite{rahman2025mgformer}         & 0.64 ± 0.18 & 0.53 ± 0.17 & 0.54 ± 0.18 & 0.54 ± 0.13  & 0.72 ± 0.15 & 0.72 ± 0.15 & 0.72 ± 0.10 & 0.67 ± 0.09  & 0.67 ± 0.13 & 0.56 ± 0.11 & 0.57 ± 0.10 &  0.57 ± 0.07 \\
\midrule
\multicolumn{13}{l}{\textbf{MoE-based Methods}} \\
\hline
GraphMoRE~\cite{guo2025graphmore}         & 0.63 ± 0.10 & 0.61 ± 0.05  & 0.52 ± 0.19  & 0.48 ± 0.16  & 0.85 ± 0.07 & 0.76 ± 0.03 & 0.53 ± 0.12 & 0.60 ± 0.10  & 0.70 ± 0.20 & 0.69 ± 0.04 & 0.49 ± 0.13 &  0.52 ± 0.15 \\
MoGE~\cite{liu2024moge}          & 0.60 ± 0.18 & 0.46 ± 0.02 & 0.39 ± 0.10  &  0.50 ± 0.00 & \underline{0.85 ± 0.03} & 0.56 ± 0.07 & 0.52 ± 0.13 & 0.52 ± 0.04 & 0.57 ± 0.15 & 0.48 ± 0.04 & 0.43 ± 0.07 &  0.50 ± 0.00 \\
Mowst~\cite{zeng2024mixture}        &  0.64 ± 0.23 & 0.59 ± 0.18 & 0.52 ± 0.11 & 0.49 ± 0.06  & 0.84 ± 0.17 & 0.78 ± 0.14 & 0.64 ± 0.09 & 0.62 ± 0.02  & 0.72 ± 0.10 & 0.63 ± 0.15 & 0.50 ± 0.12 & 0.52 ± 0.03  \\
GraphDIVE~\cite{hu2021graph}          & 0.42 ± 0.15 & 0.47 ± 0.12 & 0.53 ± 0.14 &  0.48 ± 0.09 & 0.61 ± 0.19 & 0.56 ± 0.16 & 0.68 ± 0.16 & 0.64 ± 0.09 & 0.62 ± 0.12 & 0.63 ± 0.12 & 0.58 ± 0.11 & 0.55 ± 0.07  \\
\midrule
VMoGE (w/o MGT-NFE)  & 0.57 ± 0.24 & 0.56 ± 0.16 & 0.54 ± 0.07 & 0.54 ± 0.08  & 0.73 ± 0.09 & 0.69 ± 0.08 & \underline{0.74 ± 0.08} & 0.71 ± 0.06  & 0.61 ± 0.25 & 0.60 ± 0.18 & 0.58 ± 0.12 & 0.58 ± 0.08  \\
\textbf{VMoGE (5-bands)$^{\mathrm{a}}$}
& 0.65 ± 0.04 & 0.63 ± 0.05 & \underline{0.55 ± 0.08} & 0.51 ± 0.07 & 0.79 ± 0.15 & 0.69 ± 0.11 & \textbf{0.78 ± 0.06} & \textbf{0.74 ± 0.07} & 0.61 ± 0.07 & 0.65 ± 0.17 & 0.46 ± 0.18 & 0.43 ± 0.17 \\
\textbf{VMoGE (4-bands)$^{\mathrm{b}}$} & $\mathbf{0.78 \pm 0.12}$ & $\mathbf{0.74 \pm 0.12}$ & $\mathbf{0.67 \pm 0.07}$ & $\mathbf{0.59 \pm 0.08}$ & $\mathbf{0.89 \pm 0.07}$ & $\mathbf{0.83 \pm 0.09}$ & $0.73 \pm 0.10$ & \underline{$0.72 \pm 0.09$} & $\mathbf{0.79 \pm 0.07}$ & \underline{$0.70 \pm 0.08$} & $\mathbf{0.65 \pm 0.11}$ & $\mathbf{0.61 \pm 0.13}$ \\
\midrule
Improv. & \multicolumn{1}{c}{+0.03} & \multicolumn{1}{c}{+0.02} & \multicolumn{1}{c}{+0.12} & \multicolumn{1}{c}{+0.03} & \multicolumn{1}{c}{+0.04} & \multicolumn{1}{c}{+0.03} & \multicolumn{1}{c}{+0.04} & \multicolumn{1}{c}{-} & \multicolumn{1}{c}{+0.02} & \multicolumn{1}{c}{-} & \multicolumn{1}{c}{+0.06} & \multicolumn{1}{c}{-} \\
\bottomrule
\end{tabular}
\end{adjustbox}
\begin{tablenotes}[flushleft]
\footnotesize
\item[a] Five EEG frequency bands: $\delta$ (0.5–4 Hz), $\theta$ (4–8 Hz), $\alpha$ (8–13 Hz), $\beta_{1}$ (13–25 Hz), and $\gamma$ (25–45 Hz).
\item[b] Four EEG frequency bands: $\delta$ (0.5–4 Hz), $\theta$ (4–8 Hz), $\alpha$ (8–13 Hz), and $\beta$ (13–45 Hz).
\end{tablenotes}
\end{threeparttable}
\label{table:overall}
\end{table*}

\begin{table*}[t]
\centering
\caption{Comparison of EEG-based dementia classification methods on the Open AD dataset.}
\label{tab:openad_comparison}
\renewcommand{\arraystretch}{1.2}
\setlength{\tabcolsep}{2pt} 
\begin{tabular}{p{3.0cm} p{3.0cm} p{1.3cm} p{1.5cm} p{3.0cm} p{1.8cm} p{3.6cm}}
\hline
\textbf{Study} 
& \textbf{Model type}
& \textbf{Graph-based method}
& \textbf{Task(s)} 
& \textbf{Interpretability} 
& \textbf{Datasets} 
& \textbf{Performance$^{\dagger}$} \\
\hline

Miltiadous et al. (2021)~\cite{miltiadous2021alzheimer}
& ML methods & No & AD vs FTD & Decision rules & Open AD & AD: 78.5\%, FTD: 86.3\% Acc \\

\hline
Mootoo et al. (2023)~\cite{mootoo2023detecting} 
& GDFT + SVM & Yes & AD vs HC  & Graph-frequency domain interpretation & Open AD & AD: 85\% Acc \\
\hline

Parihar et al. (2023)~\cite{parihar2023eeg} 
& Wavelet + Ensemble ML & No & AD vs HC & Physiological interpretation of EEG frequency bands & Open AD & AD: 86.9\% Acc \\

\hline
Chen et al. (2023)~\cite{chen2023multi} 
& CNN + ViT fusion & No & AD vs FTD vs NC & Attention map & Open AD & 87.33\% Acc, 88.19\% AUC \\
\hline

Lal et al. (2024)~\cite{lal2024comparative} 
& ML methods & No &  AD vs HC & Feature importance & AHEPA dataset & AD: 91\%; FTD: 93\%; AD vs FTD: 91\% Acc\\
\hline

Tawhid et al. (2025)~\cite{tawhid2025advancing} 
& CNN-based & No & AD vs HC & Layer-wise feature visualization & Open AD & AD: 97.08\%; FTD: 98.14\%  Acc\\
\hline
Siuly et al. (2025)~\cite{siuly2025investigating} 
& STFT + CNN & No & AD vs HC & Grad-CAM & Open AD & AD: 95.59\%; FTD: 95.75\% Acc\\
\hline
\hline
\textbf{VMoGE} 
& \textbf{Graph neural network and Mixture-of-Experts} 
& \textbf{Yes}
& \textbf{HC vs AD} 
& \textbf{Frequency band × Brain network × Clinical indicators} 
& \textbf{Open AD and Session-based} 
& \textbf{AD: $83\%$ Acc, $89\%$ AUC} \\
\hline
\end{tabular}
\raggedright
\footnotesize\\
$^{\dagger}$ Best performance from recent studies on the same classification task.

\label{tab:compare_methods}
\end{table*}

\subsection{Baseline Methods}

In this work, we follow the experimental settings in Table.~\ref{tab:parameters_setting} and evaluate our proposed VMoGE on two diverse datasets using five-fold cross-validation, comparing it against a wide range of baseline approaches, including traditional EEG feature extraction models, transformer-based deep learning methods, and GNN models combined with MoE mechanisms. For EEG signal classification, traditional methods include EEGNet and EEGViT, where the former employs a lightweight CNN to extract spatiotemporal and spectral features, while the latter introduces the vision transformer architecture to capture long-range dependencies in EEG sequences. In addition, methods such as DTCA-Net, Deformer, and ADFormer are representative transformer-based frameworks that focus on channel selection and multi-scale feature modeling, and have been widely applied to AD-related EEG analysis. Meanwhile, MGformer employs a multi-granularity encoder to extract temporal patterns across frequency bands, thereby enhancing the discriminative power of node representations within brain networks.

For graph-based comparative methods, we incorporated several Graph-MoE models, including GraphMoRE, MoGE, Mowst, and GraphDIVE. Additionally, we compared against recent GNN models that combine graph neural networks with expert routing strategies to effectively model the heterogeneity of brain networks and demonstrate robustness in cross-subject classification tasks.

\begin{table}
\centering
\caption{Temporal granularity configurations.}
\begin{tabular}{l l}
\hline
\textbf{Configuration} & \textbf{Granularities} \\
\hline
Fine          & [0.02, 0.03, 0.04] \\
Medium        & [0.04, 0.06, 0.08] \\
Coarse        & [0.08, 0.10, 0.12] \\
Mixed-1       & [0.02, 0.06, 0.10] \\
Mixed-2       & [0.03, 0.07, 0.11] \\
Single-Fine   & [0.02] \\
Single-Medium & [0.06] \\
Single-Coarse & [0.10] \\
\hline
\end{tabular}

\label{tab:granularity}
\end{table}

\begin{figure}
    \centering
    \includegraphics[width=0.5\textwidth]{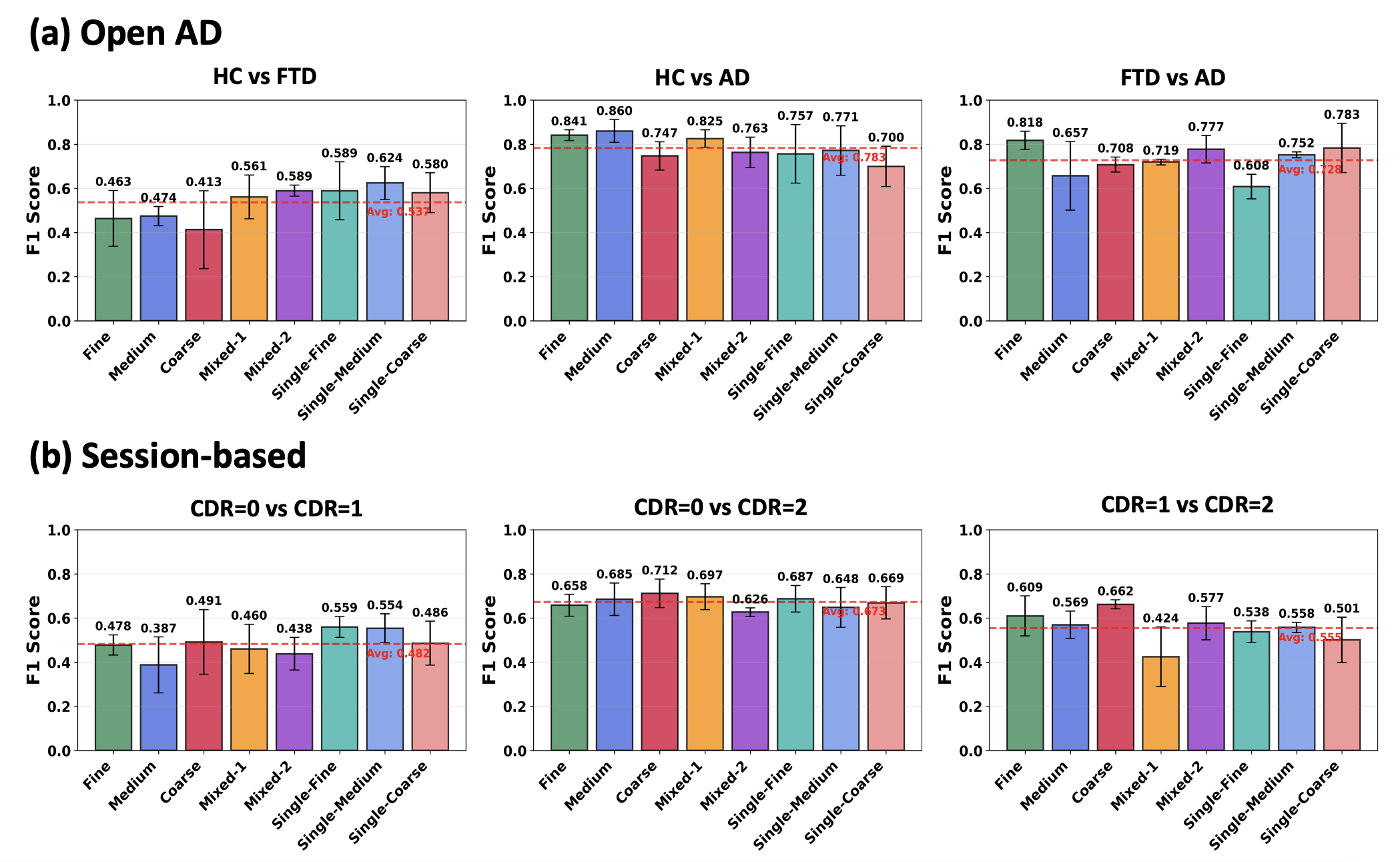}
    \caption{Ablation study of MGT-NFE module using different granularity components for EEG feature extraction.}
    \label{fig:granularities}
\end{figure} 

\begin{table*}
\centering
\caption{The $\lambda$ value ablation study results showing AUC scores across different tasks and datasets. The \textcolor{red}{\textbf{red}} values indicate the highest AUC, and the second best values are shown in \textcolor{blue}{blue} for each prior type within each task.}
\resizebox{\textwidth}{!}{
\begin{tabular}{l|ccccc|ccccc|ccccc}
\hline
\multicolumn{16}{c}{\textit{Open AD}} \\
\hline
 & \multicolumn{5}{c|}{\textbf{HC vs FTD}} & \multicolumn{5}{c|}{\textbf{HC vs AD }} & \multicolumn{5}{c}{\textbf{FTD vs AD}} \\
& $0.1$ & $0.2$ & $0.6$ & $0.8$ & $1.0$ & $0.1$ & $0.2$ & $0.6$ & $0.8$ & $1.0$ & $0.1$ & $0.2$ & $0.6$ & $0.8$ & $1.0$ \\
\midrule
No Prior & \textcolor{blue}{0.76} & 0.68 & 0.67 & 0.60 & 0.62 & \textcolor{red}{\textbf{0.83}} & \textcolor{blue}{0.78} & 0.75 & 0.70 & \textcolor{blue}{0.77} & 0.68 & 0.66 & \textcolor{blue}{0.68} & 0.61 & \textcolor{red}{\textbf{0.79}} \\
\hline
GMRF ($\textbf{L} + \lambda_{Q} \textbf{I}$) & 0.73 & \textcolor{red}{\textbf{0.72}} & 0.68 & \textcolor{blue}{0.66} & \textcolor{red}{\textbf{0.70}} & 0.77 & \textcolor{red}{\textbf{0.78}} & \textcolor{blue}{0.81} & 0.63 & 0.74 & 0.69 & \textcolor{red}{\textbf{0.75}} & \textcolor{red}{\textbf{0.76}} & \textcolor{blue}{0.71} & 0.60 \\
GMRF ($\textbf{L}{\textit{norm}} + \lambda_{Q} \textbf{I}$) & 0.74 & \textcolor{blue}{0.71} & \textcolor{red}{\textbf{0.69}} & \textcolor{red}{\textbf{0.67}} & 0.47 & \textcolor{blue}{0.80} & \textcolor{red}{\textbf{0.78}} & \textcolor{red}{\textbf{0.83}} & \textcolor{blue}{0.72} & 0.74 & \textcolor{red}{\textbf{0.74}} & \textcolor{blue}{0.70} & 0.67 & 0.68 & 0.72 \\
GMRF & \textcolor{red}{\textbf{0.77}} & 0.58 & \textcolor{blue}{0.69} & \textcolor{blue}{0.66} & \textcolor{blue}{0.65} & 0.78 & 0.77 & 0.78 & \textcolor{red}{\textbf{0.77}} & \textcolor{red}{\textbf{0.78}} & \textcolor{blue}{0.70} & 0.68 & 0.67 & \textcolor{red}{\textbf{0.84}} & \textcolor{blue}{0.74} \\
\midrule
\multicolumn{16}{c}{\textit{Session-based AD}} \\
\hline
  & \multicolumn{5}{c|}{\textbf{CDR=0 vs CDR=1}} & \multicolumn{5}{c|}{\textbf{CDR=0 vs CDR=2 }} & \multicolumn{5}{c}{\textbf{CDR=1 vs CDR=2}} \\
& $0.1$ & $0.2$ & $0.6$ & $0.8$ & $1.0$ & $0.1$ & $0.2$ & $0.6$ & $0.8$ & $1.0$ & $0.1$ & $0.2$ & $0.6$ & $0.8$ & $1.0$ \\
\midrule
No Prior & \textcolor{blue}{0.66} & 0.65 & \textcolor{red}{\textbf{0.66}} & \textcolor{blue}{0.67} & 0.61 & 0.66 & 0.73 & 0.71 & \textcolor{red}{\textbf{0.77}} & 0.68 & \textcolor{blue}{0.59} & \textcolor{red}{\textbf{0.60}} & 0.47 & 0.58 & 0.56 \\
\hline
GMRF ($\textbf{L} + \lambda_{Q} \textbf{I}$) & 0.55 & 0.61 & 0.63 & \textcolor{red}{\textbf{0.73}} & \textcolor{blue}{0.66} & 0.70 & 0.72 & \textcolor{blue}{0.71} & 0.67 & 0.64 & \textcolor{red}{\textbf{0.59}} & \textcolor{blue}{0.59} & \textcolor{blue}{0.59} & \textcolor{blue}{0.61} & \textcolor{red}{\textbf{0.62}} \\
GMRF ($\textbf{L}{\textit{norm}} + \lambda_{Q} \textbf{I}$) & \textcolor{red}{\textbf{0.69}} & \textcolor{blue}{0.65} & 0.64 & 0.60 & \textcolor{red}{\textbf{0.70}} & \textcolor{red}{\textbf{0.73}} & \textcolor{blue}{0.75} & 0.70 & \textcolor{blue}{0.70} & \textcolor{blue}{0.75} & 0.56 & 0.57 & \textcolor{red}{\textbf{0.59}} & \textcolor{red}{\textbf{0.63}} & \textcolor{blue}{0.56} \\
GMRF & 0.64 & \textcolor{red}{\textbf{0.66}} & \textcolor{blue}{0.64} & 0.61 & 0.59 & \textcolor{blue}{0.72} & \textcolor{red}{\textbf{0.76}} & \textcolor{red}{\textbf{0.81}} & 0.67 & \textcolor{red}{\textbf{0.79}} & 0.56 & 0.47 & 0.56 & 0.48 & 0.52 \\
\bottomrule
\end{tabular}
}
\label{tab:beta_ablation}
\end{table*}

\subsubsection{\textbf{Comparison With EEG Feature Extraction Models}}

Table.~\ref{table:overall} presents the VMoGE performance, including AUC and ACC, which respectively reflect discriminative capability and overall performance for EEG classification. Conventional CNN or Transformer-based EEG feature extraction models (e.g., EEGNet, EEGViT, DTCA-Net, Deformer, ADFormer, MGFormer) achieved limited performance in most tasks. For instance, in the HC vs AD setting using the Open AD dataset, Deformer and ADFormer achieved AUCs of 0.84 and 0.85, respectively, whereas our VMoGE achieved 0.89, corresponding to improvements of 0.05 and 0.04, respectively. In the HC vs FTD task, the best baseline (Deformer, AUC = 0.75) remained lower than our VMoGE (AUC = 0.78), representing an improvement of 0.03. Moreover, in the Session-based CDR=1 vs CDR=2 classification, conventional methods achieved only limited performance, with EEGNet attaining the best AUC of 0.59, whereas our VMoGE reached an AUC of 0.65, yielding an improvement of 0.06.

\subsubsection{\textbf{Comparison With Graph-MoE Models}}

Among GNN architectures incorporating MoE mechanisms (e.g., GraphMoRE, MoGE, Mowst, GraphDIVE), VMoGE achieved superior performance across all tasks on the Open AD dataset. In HC vs FTD, VMoGE attained an AUC of 0.78, surpassing the best competitor Mowst (AUC = 0.64) by 0.14. For HC vs AD, VMoGE reached an AUC of 0.89, outperforming GraphMoRE and MoGE (both AUC = 0.85) by 0.04. In the challenging FTD vs AD classification task, VMoGE led with an AUC of 0.79, exceeding Mowst (AUC = 0.72) by 0.07. These results show that conventional Graph-MoE models lack frequency-specific priors for distinguishing dementia subtypes, whereas VMoGE's GMRF-guided expert specialization consistently improves discrimination.

On the Session-based dataset, VMoGE maintained its advantage across staging tasks. In CDR=0 vs CDR=1, VMoGE achieved the highest AUC of 0.67, outperforming GraphDIVE (AUC = 0.53) by 0.14. For CDR=0 vs CDR=2, VMoGE obtained an AUC of 0.73, remaining competitive with GraphDIVE (AUC = 0.68) by 0.05. In CDR=1 vs CDR=2, VMoGE reached an AUC of 0.65, surpassing GraphDIVE (AUC = 0.58) by 0.07 and substantially exceeding Mowst (AUC = 0.50). This advantage confirms that VMoGE's structured variational inference and graph-prior regularization enable robust staging even with scarce, noisy EEG signals.

\subsubsection{\textbf{Comparison With Recent EEG-based AD Classification Studies}}

To provide a comparative perspective on VMoGE, Table.~\ref{tab:compare_methods} reports representative recent EEG-based dementia classification studies evaluated on the Open AD dataset. Although several prior works report higher single-task accuracy than VMoGE, their experimental settings differ considerably and are often limited to a single dataset, precluding a direct comparison. First, the studies listed in the Table.~\ref{tab:compare_methods} differ substantially in preprocessing pipelines, epoch length, validation protocols, and task formulation, all of which strongly affect reported performance~\cite{del2025more,miltiadous2021alzheimer}.In particular, the epoch-level cross-validation adopted in~\cite{tawhid2025advancing}  and~\cite{siuly2025investigating} can yield accuracies above 95\%, as is common with epoch-level splitting, epochs from the same subject are not separated at the subject level, potentially resulting in overly optimistic estimates of generalization performance. In contrast, the subject-independent five-fold cross-validation used in our work provides a more clinically realistic and inherently more conservative evaluation. Second, most existing methods focus exclusively on a single binary task, such as HC vs. AD. In contrast, VMoGE simultaneously addresses AD classification and three CDR-based dementia-staging tasks, thereby demonstrating broader clinical applicability. Third, our framework prioritizes interpretability and biomarker identification by linking expert gating weights to specific frequency bands, brain regions, and clinical indicators (e.g., MMSE, age, CDR severity), whereas most CNN- or wavelet-based approaches in Table.~\ref{tab:compare_methods} operate as black box predictors with limited mechanistic insight into AD or FTD neuropathology. Finally, VMoGE is validated on two independent datasets, including a clinically collected Session-based dataset with CDR staging annotations, which is rarely reported in prior studies.

\subsection{Computational Complexity for Clinical Implementation}

Following the previous experimental setting, our VMoGE demonstrates the complete advantages of MoE in terms of both training cost and inference efficiency~\cite{shazeer2017outrageously,fedus2022switch}. As shown in Table.~\ref{tab:computational_efficiency}, to ensure a fair comparison with classic transformer-based models (e.g., ADFormer, Deformer) that have higher computational complexity, we computed the latency, training time, and inference time as evaluation metrics. By exploiting expert-wise conditional computation, VMoGE avoids the heavy quadratic complexity typical of monolithic Transformer architectures and instead exhibits near-linear scaling with respect to the number of frequency-band experts. We reported that VMoGE achieved training times of 10.43–11.29s per epoch on the Open AD dataset, reducing the per-epoch training time by 2.92–3.63s compared with ADFormer (14.06-14.21s) and by 9.26–11.10 s compared with EEGDeformer (19.69-22.39s). More critically for clinical applications, VMoGE maintains remarkably low inference latency of 0.0017-0.0030s per sample, which is 2.8-6.4× faster than baseline methods. 

\begin{table}[htbp]
\centering
\caption{Training and Inference Performance Comparison Across Models and Datasets}
\label{tab:performance}
\resizebox{\columnwidth}{!}{%
\begin{tabular}{@{}llcccc@{}}
\toprule
\textbf{Task Group} & \textbf{Model} & \textbf{Dataset} & \textbf{Latency (ms)} & \textbf{Train (s)} & \textbf{Infer (s)} \\
\midrule
\multirow{6}{*}{\shortstack{HC vs FTD\\CDR=0 vs 1}} 
& \multirow{2}{*}{\textbf{VMoGE}}
  & Open AD & \textbf{$0.97 \pm 0.02$} &\textbf{$10.43 \pm 0.03$} & \textbf{$0.0017 \pm 0.0000$} \\
& & Session & \textbf{$0.93 \pm 0.01$} &\textbf{$17.66 \pm 0.12$} & \textbf{$0.0029 \pm 0.0000$} \\
\cmidrule{2-6}
& \multirow{2}{*}{ADFormer} 
  & Open AD & $0.47 \pm 0.15$ & $14.06 \pm 0.08$ & $0.0048 \pm 0.0014$ \\
& & Session & $0.48 \pm 0.06$ & $26.41 \pm 1.00$ & $0.0113 \pm 0.0012$ \\
\cmidrule{2-6}
& \multirow{2}{*}{Deformer} 
  & Open AD & $0.41 \pm 0.07$ & $19.69 \pm 3.21$ & $0.0046 \pm 0.0005$ \\
& & Session & $0.62 \pm 0.25$ & $43.73 \pm 4.80$ & $0.0146 \pm 0.0049$ \\
\midrule
\multirow{6}{*}{\shortstack{HC vs AD\\CDR=0 vs 2}} 
& \multirow{2}{*}{\textbf{VMoGE}}
  & Open AD & \textbf{$1.00 \pm 0.01$} & \textbf{$11.28 \pm 0.26$} & \textbf{$0.0018 \pm 0.0001$} \\
& & Session & \textbf{$0.97 \pm 0.01$} & \textbf{$18.70 \pm 0.13$} & \textbf{$0.0030 \pm 0.0000$} \\
\cmidrule{2-6}
& \multirow{2}{*}{ADFormer} 
  & Open AD & $0.57 \pm 0.14$ & $14.27 \pm 0.18$ & $0.0060 \pm 0.0016$ \\
& & Session & $0.49 \pm 0.01$ & $28.25 \pm 0.28$ & $0.0117 \pm 0.0004$ \\
\cmidrule{2-6}
& \multirow{2}{*}{Deformer} 
  & Open AD & $0.43 \pm 0.02$ & $21.89 \pm 0.39$ & $0.0048 \pm 0.0003$ \\
& & Session & $0.85 \pm 0.23$ & $46.90 \pm 0.95$ & $0.0191 \pm 0.0044$ \\
\midrule
\multirow{6}{*}{\shortstack{FTD vs AD\\CDR=1 vs 2}} 
& \multirow{2}{*}{\textbf{VMoGE}}
  & Open AD & \textbf{$1.01 \pm 0.00$} & \textbf{$11.29 \pm 0.13$} & \textbf{$0.0017 \pm 0.0001$} \\
& & Session & \textbf{$0.99 \pm 0.01$} & \textbf{$18.90 \pm 0.33$} & \textbf{$0.0030 \pm 0.0000$} \\
\cmidrule{2-6}
& \multirow{2}{*}{ADFormer} 
  & Open AD & $0.51 \pm 0.19$ & $14.21 \pm 0.08$ & $0.0053 \pm 0.0019$ \\
& & Session & $0.46 \pm 0.07$ & $27.77 \pm 0.39$ & $0.0110 \pm 0.0014$ \\
\cmidrule{2-6}
& \multirow{2}{*}{Deformer} 
  & Open AD & $0.56 \pm 0.24$ & $22.39 \pm 0.31$ & $0.0062 \pm 0.0029$ \\
& & Session & $0.73 \pm 0.23$ & $48.46 \pm 1.59$ & $0.0164 \pm 0.0045$ \\
\bottomrule
\end{tabular}%
}
\label{tab:computational_efficiency}
\end{table}


\subsection{Ablation study for MGT-NFE using different granularity components for EEG feature extraction.}

\subsubsection{Granularities Module Ablation}

We assessed the contribution of the MGT-NFE layer to EEG feature extraction under various temporal granularity configurations, as detailed in Table.~\ref{tab:granularity}, and evaluated classification performance across two datasets as shown in Fig.~\ref{fig:granularities}. Each configuration corresponds to a distinct temporal resolution. The \textit{Fine} setting, defined by temporal intervals of 0.02, 0.03, and 0.04 seconds, is designed to capture short-term neural events. The \textit{Medium} configuration, consisting of time windows of 0.04, 0.06, and 0.08 seconds, provides a balance between temporal stability and discriminability. In contrast, the \textit{Coarse} configuration, with intervals of 0.08, 0.10, and 0.12 seconds, emphasizes long-term rhythmic dependencies in EEG dynamics. Furthermore, the \textit{Mixed-1} and \textit{Mixed-2} configurations integrate multiple temporal scales to model cross-frequency interactions. Single-scale variants, including \textit{Single-Fine}, \textit{Single-Medium}, and \textit{Single-Coarse}, are employed as baselines to isolate the effects of individual temporal resolutions.

As shown in Fig.~\ref{fig:granularities} (a), on the Open AD dataset, the \textit{Medium} granularity achieved the highest F1-score of 0.860 in the HC vs  AD classification task, with an overall mean of 0.783. This indicates that the medium-scale temporal range of 0.04 to 0.08 seconds effectively captures stable and representative EEG dynamics associated with cognitive decline. For the HC vs ~FTD and FTD vs ~AD tasks, the \textit{Mixed-2} configuration, defined by time windows of 0.03, 0.07, and 0.11 seconds, and the \textit{Single-Fine} configuration, using a 0.02-second window, yielded the best performance (F1 $\approx$ 0.589–0.783). These results suggest that incorporating both fine- and coarse-scale temporal cues enhances cross-scale discriminability among dementia subtypes. In the Session-based dataset in Fig.~\ref{fig:granularities} (b), the \textit{Coarse} granularity, which applies time windows between 0.08 and 0.12 seconds, achieved the highest F1-score of 0.712 in the CDR=0 vs CDR=2 task. This finding indicates that longer temporal windows are more effective for capturing extended cognitive progression patterns, particularly under small-sample and high-variance conditions.


\subsubsection{Sensitivity Analysis for GMRF Priors Influence}

To investigate the impact of the regularization parameter $\lambda_{Q}$ on the prior of GMRF. We compare two GMRF prior specifications, the no-prior baseline and the structured prior defined in Eq.~\ref{eq:precision_matrix}, with analysis focusing on precision matrices of the form $\textbf{L} + \lambda_{Q} \textbf{I}$ and $\textbf{L}_{\mathrm{norm}} + \lambda_{Q} \textbf{I}$. Table.~\ref{tab:beta_ablation} presents the AUC scores for $\lambda_{Q}$ values ranging from 0.1 to 1.0 across three types of binary classification tasks on both Open and Session-based AD datasets, with red highlighting indicating the optimal $\lambda_{Q}$ value for each prior type within each task. We report that no prior GMRF achieves optimal performance at small $\lambda_{Q}$ values on the Open AD (AUC = 0.83 at $\lambda_{Q}$ = 0.1 for HC vs AD), indicating that weaker regularization benefits data-driven learning when spatial structure is not enforced. Conversely, GMRF methods exhibit distinct optimal ranges such as standard GMRF ($\textbf{L}+ \lambda_{Q} \textbf{I}$) performs best at moderate values ($\lambda_{Q}$ = 0.2 to 0.6), normalized GMRF ( $\textbf{L}_{norm} + \lambda_{Q} \textbf{I}$) achieves the highest overall score (AUC = 0.83 at $\lambda_{Q}$ = 0.6 for HC vs AD), and pure GMRF excels at both extremes, reaching the highest score with 0.84 at $\lambda_{Q}$ = 0.8 for FTD vs AD across all configurations. 

On the Session-based dataset, GMRF methods significantly improved classification performance for CDR=0 versus CDR=2 (pure GMRF: 0.81 vs no prior: 0.77) at $\lambda_{Q} = 0.6$ and $\lambda_{Q} = 0.8$. Performance progressively increased as $\lambda_{Q}$ increased from 0.1 to 0.6, suggesting that spatial regularization provides greater benefits for datasets with limited sample sizes. More broadly, the optimal $\lambda_{Q}$ values reflect the distinct characteristics of different datasets: small values (0.1-0.2) are suitable for FTD and AD with larger samples, moderate values (0.6) for balanced scenarios with clear spatial structure, and large values (0.8-1.0) for  
small-sample settings requiring stronger graph-prior regularization. 
These results indicate that spatial priors provide substantial gains when EEG brain networks exhibit inherent spatial structure.

\begin{table}[htbp]
\centering
\scriptsize
\setlength{\tabcolsep}{3pt} 
\caption{Ablation study results comparison: AUC and ACC across different tasks.}
\label{tab:ablation_results}
\resizebox{\linewidth}{!}{%
\begin{tabular}{l|cc|cc|cc}
\hline
\multicolumn{7}{c}{\textit{Open AD}} \\ \hline
\multirow{2}{*}{\textbf{Model}} & \multicolumn{2}{c|}{\textbf{HC vs FTD}} & \multicolumn{2}{c|}{\textbf{HC vs AD}} & \multicolumn{2}{c}{\textbf{FTD vs AD}} \\
 & AUC & ACC & AUC & ACC & AUC & ACC \\ \hline

S-E ($\delta$) & \textbf{0.78±0.16} & 0.63±0.17 & 0.85±0.14 & 0.65±0.17 & 0.78±0.18 & 0.54±0.12 \\
S-E ($\theta$) & 0.69±0.17 & 0.57±0.09 & 0.89±0.10 & 0.73±0.21 & 0.85±0.16 & 0.63±0.10 \\
S-E ($\alpha$) & 0.74±0.09 & 0.54±0.14 & 0.80±0.18 & 0.57±0.13 & 0.79±0.20 & \textbf{0.74±0.15} \\
S-E ($\beta$)  & 0.67±0.20 & 0.54±0.09 & 0.76±0.16 & 0.60±0.16 & 0.74±0.13 & 0.52±0.13 \\
\textbf{VMoGE}        & 0.77±0.08 & \textbf{0.65±0.15} & \textbf{0.92±0.10} & \textbf{0.78±0.12} & \textbf{0.87±0.12} & 0.70±0.08 \\ \hline

\multicolumn{7}{c}{\textit{Session-based}} \\ \hline
\multirow{2}{*}{\textbf{Model}} & \multicolumn{2}{c|}{\textbf{CDR=0 vs CDR=1}} & \multicolumn{2}{c|}{\textbf{CDR=0 vs CDR=2}} & \multicolumn{2}{c}{\textbf{CDR=1 vs CDR=2}} \\
 & AUC & ACC & AUC & ACC & AUC & ACC \\ \hline

S-E ($\delta$) & 0.66±0.11 & 0.58±0.07 & 0.78±0.11 & 0.57±0.09 & 0.68±0.10 & 0.58±0.06 \\
S-E ($\theta$) & 0.59±0.08 & 0.47±0.06 & 0.78±0.06 & 0.64±0.08 & 0.52±0.08 & 0.54±0.08 \\
S-E ($\alpha$) & 0.66±0.06 & 0.56±0.05 & \textbf{0.82±0.06} & \textbf{0.68±0.11} & 0.63±0.09 & 0.50±0.00 \\
S-E ($\beta$)  & 0.64±0.10 & 0.51±0.07 & 0.74±0.16 & 0.53±0.10 & 0.55±0.12 & 0.54±0.12 \\
\textbf{VMoGE}        & \textbf{0.69±0.11} & \textbf{0.60±0.16} & 0.72±0.09 & 0.62±0.12 & \textbf{0.68±0.09} & \textbf{0.60±0.07} \\ \hline
\end{tabular}%
}
\raggedright
\footnotesize S-E: Single-Expert using individual frequency band networks. 
\label{table:MoE_comparision}
\end{table}

 \begin{figure}
    \centering
    \includegraphics[width=0.5\textwidth]{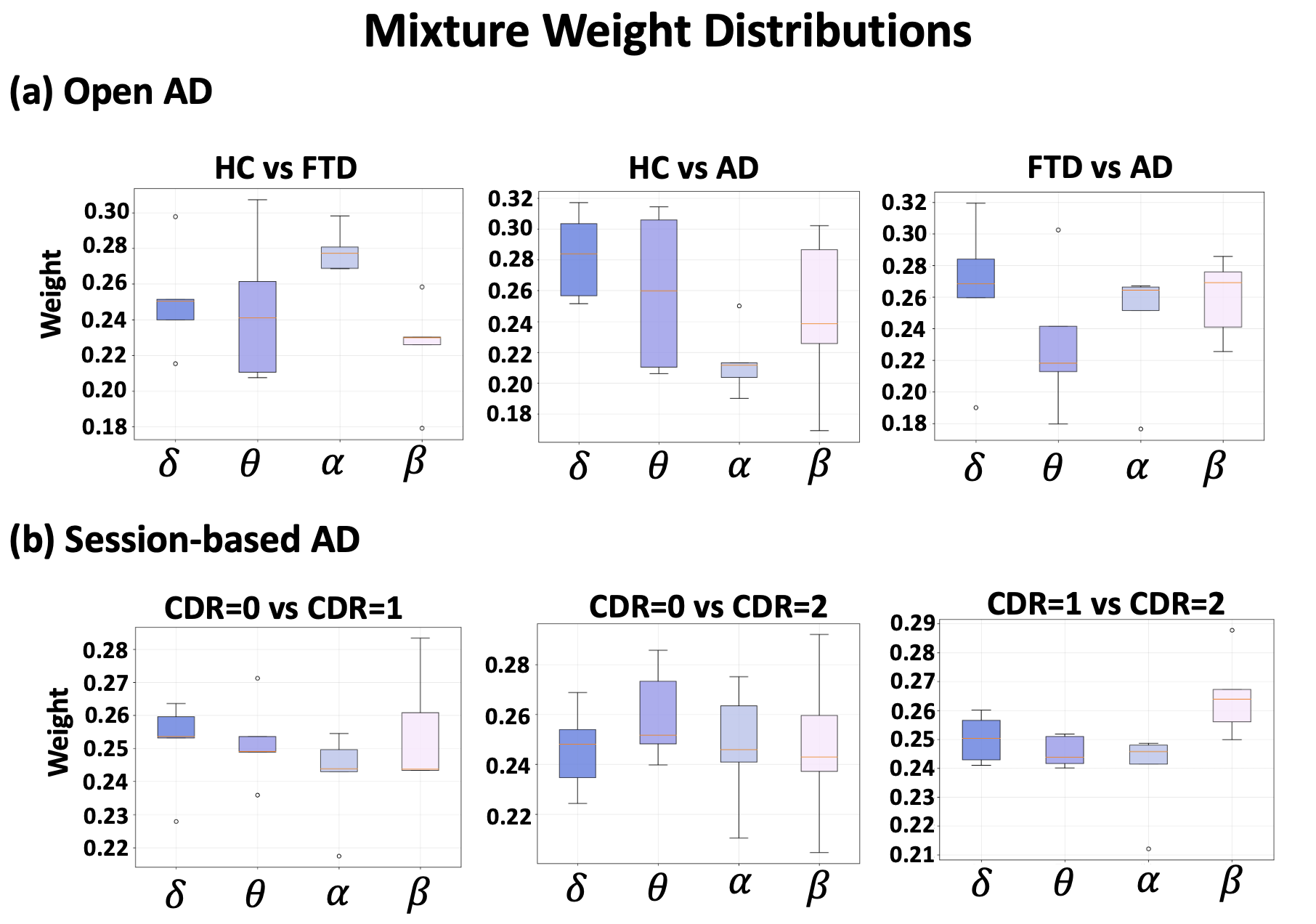}
    \caption{Box plots showing the distribution of mixture weights across different bands under three subtyping comparison conditions. The weights represent the contribution of different components to distinguishing between HC, FTD, and AD in the Open AD dataset in Fig. (a), and to Session-based AD staging based on CDR=0, CDR=1, and CDR=2 in Fig. (b).}
    \label{fig:VMoE_Weight}
\end{figure}

\begin{figure*}
    \centering
    \includegraphics[width=1\textwidth]{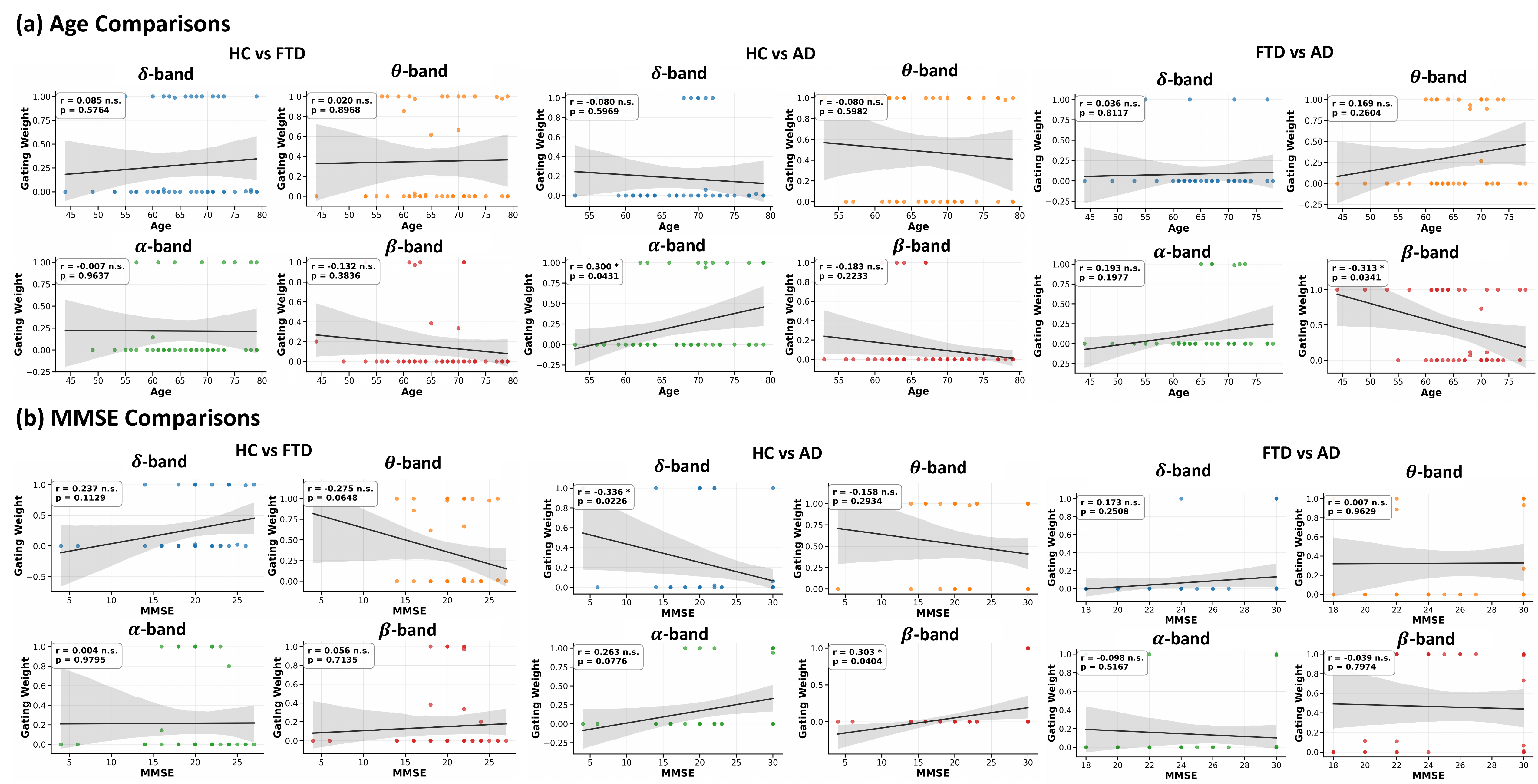}
    \caption{The expert gating weight analysis for different dementia indicator analysis. The top row shows age-based comparisons, and the bottom row shows MMSE-based comparisons. Each column represents a different pairwise comparison: HC vs. FTD, HC vs. AD, and FTD vs. AD.}
    \label{fig:gating_scatter_comparison}
\end{figure*}

\subsubsection{Single-Expert and Full-Expert Ablation}

In this section, we observe that flexible prior gating facilitates optimal allocation of frequency band expert outputs. In the experiments presented in Table.~\ref{table:MoE_comparision}, we compare the Single-Expert (S-E) model, which uses only a single frequency band expert, with the Full-Expert (VMoGE) model, which integrates all frequency bands. In the Open AD, the performance of S-E models varied substantially across frequency bands. For example, $\delta$ (AUC = 0.85) and $\theta$ (AUC = 0.89) showed advantages in the HC vs AD task, while $\alpha$ achieved the highest accuracy in the FTD vs AD task (ACC = 0.74). In contrast, $\beta$ generally underperformed, suggesting that high-frequency activity alone is insufficient for robust classification. In the Session-based dataset, the $\alpha$ band performed best in CDR=0 vs CDR=2 (AUC = 0.82), but overall accuracies were low and task-dependent fluctuations indicated the limitations of single-band experts in heterogeneous clinical data.

By contrast, VMoGE demonstrated overall stable and competitive performance across both datasets. In the Open AD dataset, it achieved the best results in HC vs AD (AUC = 0.92, ACC = 0.78) and maintained strong performance in FTD vs AD (AUC = 0.87). On the Session-based dataset, VMoGE achieved balanced results across the three CDR staging tasks, with AUCs of 0.69 (CDR=0 vs 1) and 0.68 (CDR=1 vs 2), demonstrating robustness against heterogeneous and limited-sample EEG settings.

\begin{figure}
    \centering
    \includegraphics[width=0.5\textwidth]{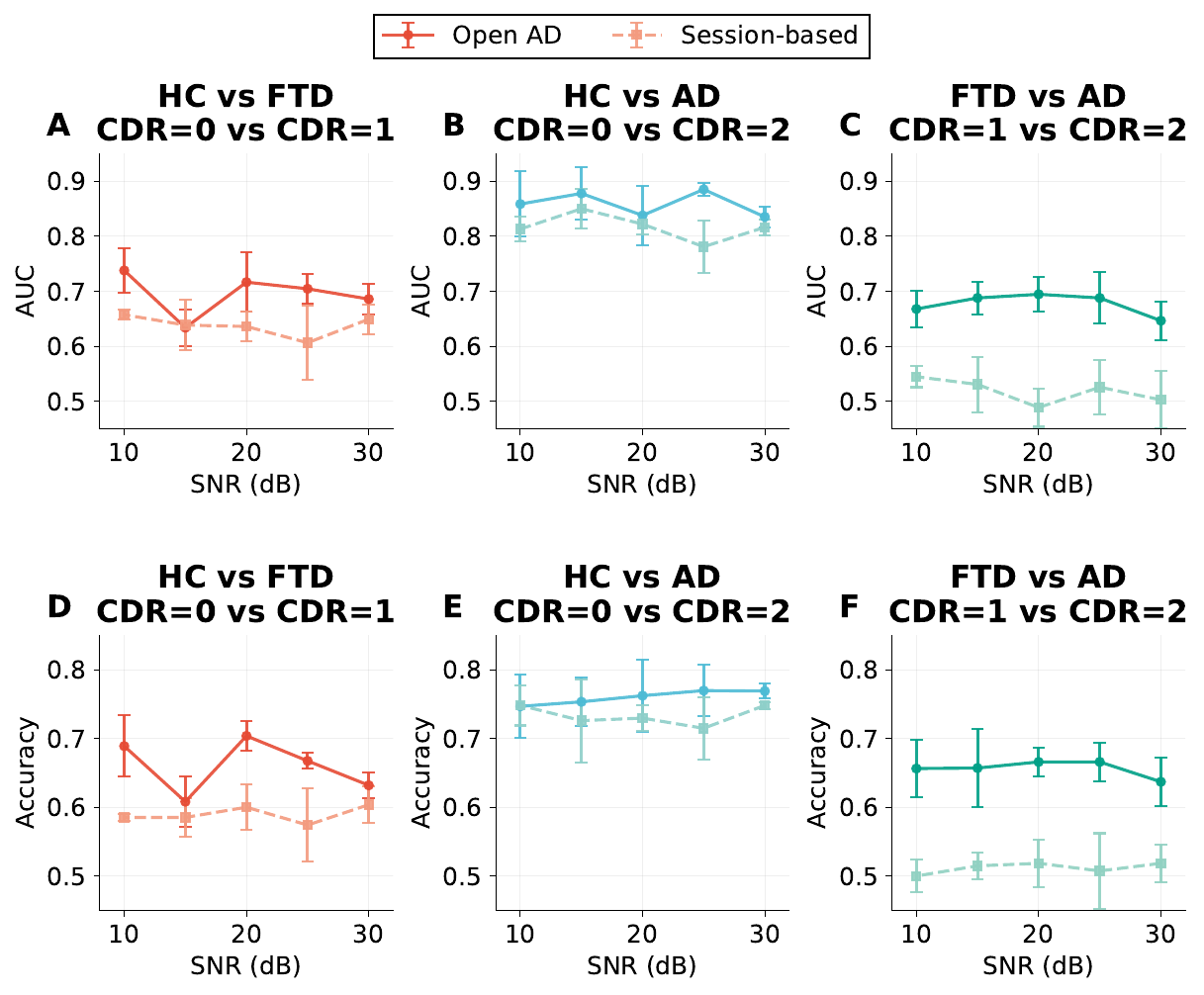}
    \caption{Robustness evaluation of VMoGE under additive white Gaussian noise across SNR levels of 10, 20, and 30 dB. Panels A–C and D–F report AUC and accuracy, respectively, for three classification tasks on the Open AD and Session-based datasets. }
    \label{fig:noisy_analysis}
    \vspace{-3mm}
\end{figure}

\begin{figure}
    \centering
    \includegraphics[width=0.43\textwidth]{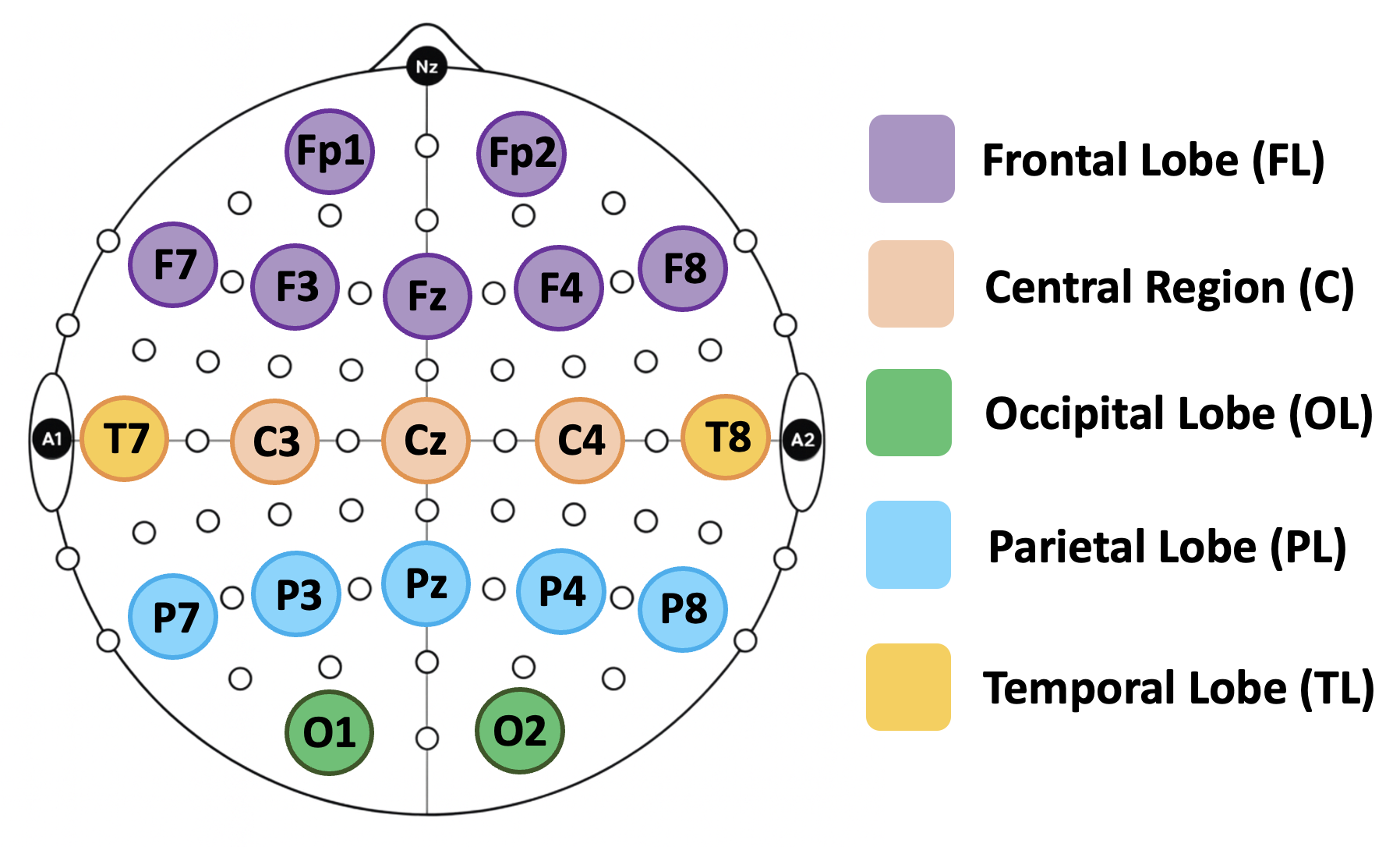}
    \caption{Diagram of electrode spatial positions and corresponding brain regions in the EEG 10–20 system~\cite{sanchis2024novel}, where different colors represent distinct cortical areas.}
    \label{fig:7_group}
    \vspace{-3mm}
\end{figure}

\begin{figure*}
    \centering
    \includegraphics[width=0.9\textwidth]{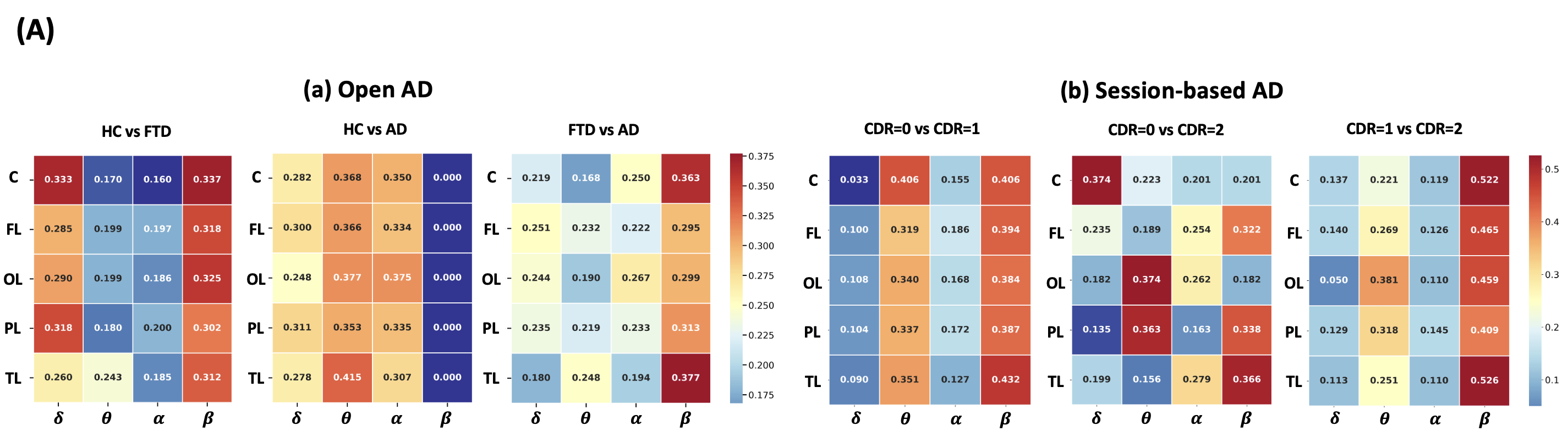}

    \vspace{0.5cm}
    \includegraphics[width=0.9\textwidth]{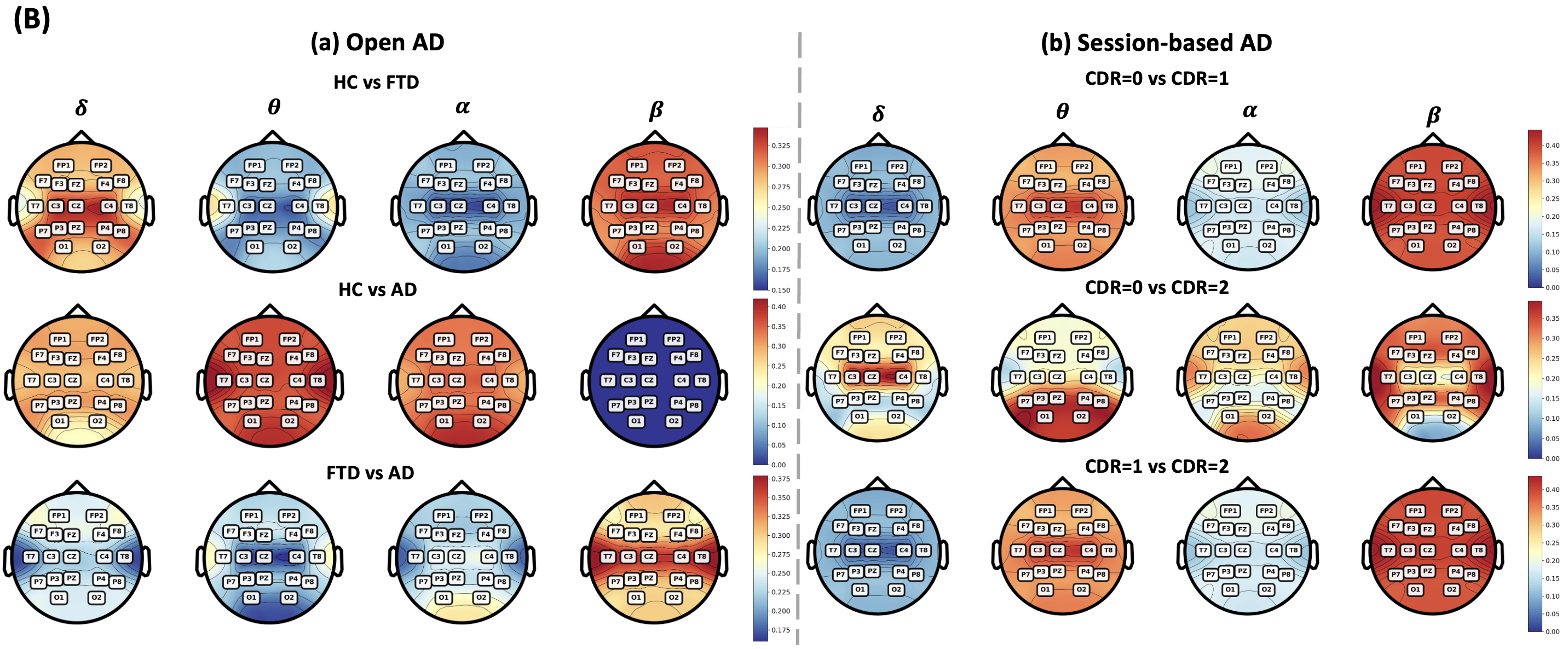}

    \caption{Combined visualization of heatmaps (upper) and topographical maps (lower), showing spatial distributions and activation patterns across different conditions.}
    \label{fig:combined}
    \vspace{-1mm}
\end{figure*}

\section{Noise in EEG Signal Analysis}\label{sec:noise_analysis}

Since EEG signals are inherently susceptible to noise interference, we introduced a framework with graph priors and evaluated its robustness under realistic clinical conditions by incorporating additive white Gaussian noise at signal-to-noise ratio (SNR) levels of 10, 20, and 30 dB. As shown in Fig.~\ref{fig:noisy_analysis}, the Open AD dataset (solid lines) consistently outperformed the Session-based dataset (dashed lines) across all SNR levels, reflecting its higher signal quality and larger sample size. In the HC vs. AD task (panels B and E), both datasets maintained relatively stable performance across the tested SNR range, with Open AD sustaining AUC values between 0.83 and 0.89, indicating that the discriminative contrast between healthy controls and AD patients is sufficiently robust against moderate noise levels. For the HC vs. FTD task (panels A and D), Open AD showed moderate sensitivity to noise, with AUCs ranging from 0.63 to 0.75. In contrast, Session-based AD exhibited lower and more variable performance, suggesting that $\alpha$-band features distinguishing FTD from healthy controls are more susceptible to noise contamination.
Notably, VMoGE maintained stable performance as SNR increased on the Open AD dataset without consistent monotonic degradation, suggesting that the GMRF-based structural priors contribute to noise resilience by enforcing graph-smoothness constraints that suppress spurious perturbations. These results demonstrate that VMoGE can sustain clinically acceptable performance in complex, realistic clinical environments.

\vspace{-2mm}
\section{Model Interpretability}\label{sec:explainable_diagnosis}

\subsection{Gating Weights for Biomarker Analysis}

In Fig.~\ref{fig:VMoE_Weight}, we further examine the dynamic allocation of expert weights across frequency bands in the VMoGE model. For cross-group discrimination in Open AD tasks, the model assigned markedly higher weights to the $\delta$ and $\theta$-bands in the HC vs AD group, while reducing the contribution of the $\alpha$- and $\beta$-bands. This aligns with the well-established pathological slowing in AD, characterized by enhanced slow-wave ($\delta\slash\theta$) activity and attenuated $\alpha\slash\beta$ rhythms. In contrast, for the HC vs FTD group, the $\alpha$-band dominated the gating weights that highlight the diagnostic relevance of $\alpha$-rhythm alterations in distinguishing frontotemporal dementia from normal aging. When comparing FTD vs AD, the gating weights were more balanced, with $\delta$, $\beta$, and $\alpha$ receiving relatively higher contributions, indicating that a combination of slow-wave and high-frequency abnormalities best captures the differences between these two dementia subtypes.

For disease progression analysis in Session-based AD tasks, the weight distribution demonstrated a different pattern. As the disease progressed to CDR=2, the $\theta$ and $\alpha$-bands exhibited further weight increases, reflecting the progressive prominence of slow-wave activity, while these bands displayed greater variability, indicating that, in some individuals, high-frequency abnormalities remained pronounced. Overall, as CDR increased from 0 to 2, the model gradually shifted from a balanced reliance on multiple bands to an emphasis on slow-wave ($\delta/\theta$) signals, consistent with the clinical trajectory from mild to severe dementia.

\subsection{Effects of Cognitive Function and Age on Expert Weight Allocation}

In this section, we further investigate the interpretability of the VMoGE in AD diagnosis by analyzing the associations between the gating weights of frequency band-specific experts and clinical variables (e.g., age and MMSE scores) as presented in Fig.~\ref{fig:gating_scatter_comparison}. In analyzing the Open AD dataset and the influence of age on expert weight allocation, we found that no frequency bands showed significant correlations between the HC and FTD groups, indicating that the EEG features learned by the model are relatively stable and unaffected by age. In the HC vs AD comparison, the $\alpha$-band showed higher diagnostic value in elderly subjects, as evidenced by increased discriminative weights assigned by the model ($r = 0.300$, $p = 0.0431$). This likely reflects the diagnostic contrast between preserved $\alpha$ characteristics in normal aging and AD-related pathological changes (e.g., $\alpha$ slowing and desynchronization). In FTD vs AD, the $\beta$-band exhibited a significant negative correlation ($r = -0.313$, $p = 0.0341$), indicating that it possesses greater discriminative capacity in younger patients.

To better measure cognitive function using MMSE scores, we observed that in the HC vs AD task, the $\delta$-band exhibited a significant negative correlation ($r = -0.336$, $p = 0.0226$), indicating that patients with poorer cognitive function showed greater discriminative value in this frequency band. This finding is consistent with the pathological increase in $\delta$ waves observed in AD patients, suggesting that $\delta$ waves become a more prominent neurophysiological marker in the late stages of the disease. In contrast, the $\beta$-band demonstrated a potential value in early diagnosis ($r = 0.303$, $p = 0.0404$), suggesting that patients with better cognitive function retained more $\beta$ wave characteristics.

\subsection{Gating Weights for Spatial Brain Attribute Activation Analysis}

Following prior research~\cite{acharya2016american,sanchis2024novel}, we aligned the channel clustering of electrodes' spatial regions across major cortical regions, including Frontal Lobe (FL), Central Region (C), Occipital Lobe (OL), Parietal Lobe (PL), and Temporal Lobe (TL), as shown in Fig.~\ref{fig:7_group}. This alignment enables us to analyze and compare the spatial gating weights of different bands according to their spatial heatmaps, as illustrated in Fig.~\ref{fig:combined} (A) and topographical maps in Fig.~\ref{fig:combined} (B). In the Open AD dataset, the HC vs.\ AD comparison showed elevated $\theta$- and $\alpha$-band spatial weights over posterior regions (OL: $\theta = 0.377$, $\alpha = 0.375$; PL: $\theta = 0.353$, $\alpha = 0.335$), with the highest regional weight observed for the $\theta$ band in the temporal lobe ($0.415$), consistent with parieto-occipital network involvement in AD. For FTD vs AD, the $\beta$-band dominated the weight allocation and was concentrated in central and occipital regions (C3, Cz, C4, O1, O2), revealing the diagnostic value of high-frequency oscillatory abnormalities.

In the Session-based AD, disease progression from CDR=0 to CDR=2 was accompanied by the spatial expansion of brain region activation. For CDR=0 vs CDR=1, the $\beta$ band received the highest weights across most regions, particularly in the temporal lobe (TL: $0.432$) and central region (C: $0.406$). For CDR=0 vs CDR=2, the weighting shifted toward slow-wave activity, with the $\delta$ band dominating the central region ($0.374$) and the $\theta$ band dominating posterior regions (OL: $0.374$; PL: $0.363$). For CDR=1 vs CDR=2, the $\beta$ band again dominated, with the largest weights observed in the temporal and central regions (TL: $0.526$; C: $0.522$), while the $\theta$ band remained moderately weighted over the posterior region (OL: $0.381$). These findings suggest that slow-wave features are most informative for distinguishing normal cognition from severe dementia, whereas central--temporal $\beta$-band activity contributes more strongly to discriminating adjacent severity stages.

\vspace{-2mm}
\section{Discussion}\label{sec:discussion}

In this section, we further discuss VMoGE across three dimensions: (1) frequency-specific biomarkers in EEG that reveal distinct pathophysiological signatures, (2) spatial brain network alterations, and (3) structured graph modeling. The gating weight analysis in Fig.~\ref{fig:VMoE_Weight} shows that VMoGE primarily emphasized slow-wave activity for HC vs. AD classification, especially the $\delta$- and $\theta$-band contributions. This finding is consistent with AD-related EEG slowing~\cite{kopvcanova2024resting,benz2014slowing}, in which slow-wave activity becomes more prominent as cognitive impairment progresses. As shown in the age and cognitive function analysis in Fig.~\ref{fig:gating_scatter_comparison}, the significant negative association between $\delta$-band weights and MMSE scores ($r = -0.336$, $p = 0.0226$) further supports the role of $\delta$ activity as a marker of more severe cognitive decline: as the condition worsens and MMSE scores decline~\cite{kwak2006quantitative}. In contrast, $\beta$ power is positively related to cognitive performance ($r = 0.303$, $p = 0.0404$), emphasizing its importance in cognitive processing~\cite{azami2023beta,torabinikjeh2022correlations} and its contribution to working memory, executive control, and motor regulation~\cite{guntekin2013beta,klados2016beta,schmidt2019beta}. Notably, the $\beta$-band demonstrated stronger discriminative capacity in younger patients ($r = -0.313$, $p = 0.0341$), pointing to its potential as a biomarker for early-onset dementia in FTD vs AD.

At the spatial level, Fig.~\ref{fig:combined} shows that the $\theta$- and $\alpha$-band weights were elevated over posterior regions for HC vs.\ AD classification, consistent with disrupted posterior rhythms and impaired parietal--occipital cognitive networks~\cite{choong2025evaluating,baik2022implication}. In the Session-based AD dataset, spatial weight allocation varied across disease stages rather than following a monotonic pattern~\cite{kudo2024neurophysiological}. Slow-wave contributions were most prominent when distinguishing normal cognition from severe dementia (CDR=0 vs CDR=2), with the $\delta$ band dominating the central region and the $\theta$ band dominating posterior regions~\cite{kwak2006quantitative}. In contrast, adjacent-stage discrimination in CDR=1 vs CDR=2 relied primarily on $\beta$-band weights over the central (C3, Cz, and C4) and temporal (T7 and T8) regions. These findings indicate that $\beta$-band alterations are region-specific rather than uniformly attenuated and provide staging-relevant information that complements slow-wave markers~\cite{smailovic2019neurophysiological,iqbal2026quantitative}.

By integrating frequency-specific GMRF priors into the variational inference framework, this study successfully captured the structured dependencies of EEG brain networks. Ablation experiments shown in the Table.~\ref{tab:beta_ablation} demonstrated that GMRF priors yielded significant performance improvements on the sample-limited session-based dataset, with AUC increasing from $0.77$ (CDR=0 vs CDR=2, $\lambda_{Q}=0.8$) to $0.81$ (CDR=0 vs CDR=2, $\lambda_{Q}=0.6$). This confirmed the regularization benefits of graph-structured priors in small-sample scenarios. The task-dependent optimal $\lambda_{Q}$ values also reflected the heterogeneity of spatial smoothness in brain networks across different pathological states, with FTD/AD tasks favoring $\lambda_{Q}$ ranging from $0.2$ to $0.8$, while CDR=1/CDR=2 staging tasks favored $\lambda_{Q}$ ranging from $0.6$ to $0.8$.

While VMoGE achieves strong subtype classification and severity staging from EEG alone, the multi-level neurodegenerative processes of AD are difficult to capture through a single modality. Recent work shows that fusing EEG with structural MRI via attention improves AD classification~\cite{liu2025multimodal,paduvilan2025attention,chen2025toward}, that combining neuroimaging with genomic single-nucleotide polymorphism  data captures complementary disease signatures~\cite{venugopalan2021multimodal}, that gene–brain interaction graph priors enhance both accuracy and interpretability~\cite{zhang2025deg}, and that supervised contrastive learning across SNPs, proteomics, and MRI mitigates modality heterogeneity~\cite{xie2026multi}. VMoGE is naturally suited to such extensions: each expert could represent a distinct modality (e.g., fMRI functional connectivity or genotype-informed networks), with the variational gating mechanism dynamically weighting modality-specific contributions, and the GMRF prior generalized to encode cross-modal dependencies such as structure–function coupling from simultaneous EEG-fMRI or gene–brain networks~\cite{sun2024structure}. This multimodal extensibility positions VMoGE to enable more precise biomarker identification, more sensitive early diagnosis, and personalized disease risk stratification.

\section{Conclusions}\label{sec:conclusion}
In this paper, we propose the VMoGE framework, a variational mixture of graph neural experts framework that integrates an MGT-NFE extractor capable of capturing multi-scale EEG features with GMRF-based structural graph priors. Through structured variational inference, the model dynamically allocates expert weights, enabling adaptive integration of multi-frequency band information to achieve a precise dementia diagnosis based on frequency-specific EEG biomarkers. Furthermore, VMoGE reveals critical neurophysiological insights across two AD diagnostic datasets. These frequency-specific and spatially-localized biomarkers not only align closely with established neuropathological mechanisms but also demonstrate the effectiveness of the structured variational inference framework in capturing brain network heterogeneity. This work provides a deep learning solution that combines both accuracy and clinical interpretability for early dementia screening and disease subtype differentiation.


\bibliographystyle{ieeetr}
\bibliography{IEEEtranTSP}

\end{document}